\documentclass[]{article}
\usepackage[a4paper, total={7in, 9in}]{geometry}
\usepackage{url}
\usepackage{cite}
\usepackage{graphicx}
\usepackage[cmex10]{amsmath}
\usepackage{amssymb}
\usepackage{array}
\usepackage{fixltx2e}
\usepackage{tikz}
\usepackage{algpseudocode}
\usepackage{flushend}
\usepackage{booktabs}
\usepackage{mathtools}
\usepackage{stmaryrd}
\usepackage{subcaption}

\begin{document}
	
	\title{Addressing Multilabel Imbalance with an Efficiency-Focused Approach Using Diffusion Model-Generated Synthetic Samples}
	\author{F. Charte\thanks{Corresponding author: Francisco Charte Ojeda, Computer Science Department, A3-219 Universidad de Ja\'en, Campus Las Lagunillas, 23071 Ja\'en, Spain.\\ Francisco Charte, fcharte@ujaen.es. Miguel Ángel Dávila, mdavila@ujaen.es. María Dolores Pérez-Godoy, lperez@ujaen.es. María José del Jesus, mjjesus@ujaen.es.} \and
		M. A. D\'avila \and
		M. D. P\'erez-Godoy \and
		M. J. del Jesus}

\maketitle

\begin{abstract}
	Predictive models trained on imbalanced data tend to produce biased results. This problem is exacerbated when there is not just one output label, but a set of them. This is the case for multilabel learning (MLL) algorithms used to classify patterns, rank labels, or learn the distribution of outputs. Many solutions have been proposed in the literature. The one that can be applied universally, independent of the algorithm used to build the model, is data resampling. The generation of new instances associated with minority labels, so that empty areas of the feature space are filled, helps to improve the obtained models. The quality of these new instances depends on the algorithm used to generate them. In this paper, a diffusion model tailored to produce new instances for MLL data, called MLDM (\textit{MultiLabel Diffusion Model}), is proposed. Diffusion models have been mainly used to generate artificial images and videos. Our proposed MLDM is based on this type of models. The experiments conducted compare MLDM with several other MLL resampling algorithms. The results show that MLDM is competitive while it improves efficiency.
\end{abstract}
	
\section{Introduction}\label{Sec.Introduction}
	{T}{raining} a machine learning (ML) model with imbalanced data~\cite{he2009learning} typically introduces bias into the outputs it produces. The fewer samples there are for a given class during training, the greater the likelihood that the model will make incorrect predictions. The challenges machine learning algorithms face in dealing with this problem have been widely studied and documented~\cite{kaur2019A}.
	
	Although this is a problem that exists in different types of data and affects disparate types of ML models, the present work focuses on a particularly difficult case, namely multilabel learning (MLL). This includes tasks such as multilabel classification (MLC)~\cite{CharteMLC}, multilabel ranking~\cite{vembu2010label} and label distribution learning~\cite{geng2016label}. All of them have to work with data presenting high imbalance levels~\cite{charte2015addressing}, as well as other difficulties related to the multiple label nature of the data~\cite{charte2019dealing}. 
	
	Several ways to handle imbalanced data while training machine learning models can be found in the literature. These methods fall into three main groups:

	\begin{itemize}
		\item \textbf{Algorithm-level tweaks}: involve changing existing methods to work better with minority classes. For example, we might adjust decision thresholds~\cite{maloof2003learning} or calibrate levels of confidence.

		\item \textbf{Cost-sensitive learning}: means making the model pay more attention to minority classes by assigning higher costs to their misclassification. This can be done through cost-sensitive boosting or by weighting classes differently during training~\cite{sun2007cost}.
		
		\item \textbf{Classifier Ensemble}: addresses the imbalance problem by using ensembles of models~\cite{liu2009easyensemble}, which are more robust to this problem, instead of a single classifier.

		\item \textbf{Data resampling}: tries to balance out the classes in the training data. This can be achieved by oversampling minority classes (the SMOTE approach), undersampling majority classes, or a combination of the two.
	\end{itemize}

	Each approach has its own set of advantages and disadvantages, and the optimal choice depends on the specific problem, dataset, and learning algorithm being used. Nevertheless, data resampling is a widely utilized approach, as it can be employed as a preprocessing step preceding the implementation of any machine learning model.

	Data resampling is a common way to fix data problems in machine learning. Oversampling is usually preferred over undersampling because it preserves and enhances the available information. Chawla et al.~\cite{Chawla2002} introduced SMOTE, a widely used method that generates synthetic minority instances, effectively increasing the representation of rare classes without discarding valuable data. Undersampling techniques remove instances from the majority class, which can result in the loss of crucial information~\cite{Liu2009}. He and Garcia review various resampling methods~\cite{he2009learning}, emphasizing the advantages of oversampling. Oversampling is preferred because it adds data without losing existing samples. This helps the model learn from all available information. Recently, oversampling techniques more advanced than those based on nearest neighbors (NN), such as Gaussian mixture models, have appeared.
	
	Oversampling methods are typically preferred in MLL as well, given that removing a sample from an MLD implies a greater loss of information than in standard datasets, due to the specific traits of these datasets (a more detailed examination of this topic will be provided below). One significant limitation of many of these algorithms, including those founded in SMOTE, is their efficiency. The process of searching for NNs, taking into account the high set of features of many MLDs, and sometimes also a vast set of labels, is both time-consuming and resource-intensive.

	Modern generative models, such as (VAEs)~\cite{kingma2013auto}, generative adversarial networks (GANs)~\cite{goodfellow2014generative}, normalizing flows (NMs)~\cite{rezende2015variational}, denoising diffusion probabilistic models (DDPMs)~\cite{sohl2015deep}, and transformers~\cite{vaswani2017attention}, have shown great ability to produce synthetic data, and are being mostly used to generate fictional images and text generation. Among the existing approaches, diffusion models have shown great advantages, as will be discussed later in section~\ref{Sec.DiffusionMOdels}.
		
	The method we introduce in this work, designated as MLDM (\textit{MultiLabel Diffusion Model}), has been devised with three principal objectives in mind: 
	\begin{itemize}
		\item The first objective is to generate high-quality synthetic instances by following the approach of existing multilabel oversampling algorithms (MOA), such as MLSMOTE~\cite{mlsmote} or MLSOL~\cite{mlsol}, rather than simply duplicating samples with minority labels.
		
		\item The second one is to consistently reduce the imbalance level of MLDs while avoiding the increase in imbalance that some MOA produce.
		
		\item Lastly, the preceding two tasks should be completed in an efficient manner, with a reduction in both time and resource consumption when compared to a similar MOA. 
	\end{itemize}

	The roadmap for this study unfolds as follows: We begin by laying the groundwork in Section~\ref{Sec.RelateWork}, where we delve into the key concepts of MLL, imbalance issues, and diffusion models that underpin our research. Section~\ref{Sec.MLDM} unveils our novel MLDM algorithm, offering an in-depth exploration of its inner workings. The experimental setup and results are provided in Section~\ref{Sec.Experiments}, encompassing the chosen datasets, performance metrics, and implementation specifics, alongside the obtained outcomes. A critical analysis of these findings takes center stage in Section~\ref{Sec.Discussion}, where we benchmark MLDM against existing methodologies. To wrap up, Section~\ref{Sec.Conclusions} distills the essence of our investigation, highlighting the key takeaways and their implications for the field.
	
\section{Related work}\label{Sec.RelateWork}

This section provides the necessary background to contextualize our proposal. The following subsection provides an overview of the multilabel learning task. Subsequently, the challenge of learning from imbalanced data and the distinctive characteristics of imbalanced MLDs are elucidated. Finally, the foundation of diffusion models is outlined.

\subsection{Multilabel Learning}

Supervised machine learning models are trained on labeled data, meaning that each combination of features is assigned an output value. This can be continuous or discrete, resulting in regression or classification tasks. In the latter case, a single label is usually assigned to each data instance. However, real-world scenarios can be more complex than binary or multiclass classification problems. One such scenario is multilabel learning (MLL)~\cite{CharteMLC}, i.e. each data point has multiple labels.

In MLL, we have a dataset with features $X^1,\dots X^f$ and a set of distinct labels $L$. Each data point $I_i$ in the dataset is defined as $(X_i, Y_i)$, where $X_i$ is a feature vector and $Y_i$ is a subset of $L$. This means that each data point can have multiple labels. The goal of MLL is to find a model that can predict the subset of labels applicable to each instance. In other words, we want to find a function $f(x_i)$ that maps each feature vector $x_i$ to a subset of labels $Y'_i \in L$.

\begin{equation}\footnotesize
	I_i = (X_i, Y_i) \mid X_i \in X^1\times X^2\times \dots\times X^f, Y_i \subseteq L .
	\label{Eq.MLL}
\end{equation}
\begin{equation} \footnotesize
	Y'_i=f(x_i), Y'_i \in L  . 
	\label{Eq.MLLf} 
\end{equation}

MLL has been applied in various fields, including aviation safety~\cite{robinson2018multi}, remote sensing image retrieval~\cite{dai2018novel}, chemical toxicity prediction~\cite{liu2018integrated}, and automatic post tagging~\cite{stackexchess}. Several reviews of these works and other MLL techniques are available~\cite{Zhang:2013,han2023survey}.

The degree of \textit{multilabelness} in a dataset $D$, defined as the number of labels present in each of its samples, is typically quantified using two established metrics. The label cardinality (\ref{Eq.CardDens}) is defined as the mean number of active labels in an MLD. It seems reasonable to posit that a larger set of labels $L$ will also result in a higher cardinality. For this reason, it is common practice to also compute label density (\ref{Eq.CardDens}). These two metrics, along with others such as the total number of labels in the MLD and the {MeanIR} defined below, are typically provided to characterize MLDs in experiments.

\begin {equation}\footnotesize
\begin{aligned}[c]
\mathit{Card}(D) =\frac{1}{|D|} \sum_{i=1}^{|D|} |Y_i|
\end{aligned}
\qquad
%\end {equation}
\begin{aligned}[c]
%\begin {equation}\footnotesize
\mathit{Dens}(D) =\frac{1}{|L|} \frac{1}{|D|} \sum_{i=1}^{|D|} |Y_i|
\label {Eq.CardDens}
\end{aligned}
\end {equation}

\subsection{Learning from Imbalanced Data}

Often, the data used to train classifiers is imbalanced. This means that some class labels have more instances than others. To measure this imbalance, we use the imbalance ratio (IR), which is defined as the number of samples in the majority class divided by the number of samples in the minority class. Metrics other than IR have been proposed in the literature, but they have not been adapted to the multilabel case.

In binary scenarios, the IR is computed directly, as there are only two labels, one representing the majority and the other representing the minority. In multiclass scenarios, individual IRs are calculated pairwise and then averaged to obtain the mean IR~\cite{Alberto:2013}. A similar procedure is followed in MLL (\ref{Eq.MeanIR}) to obtain the {MeanIR} metric introduced in~\cite{charte2015addressing}.

\begin {equation}\footnotesize
\mathit{MeanIR} = \frac{1}{\mid Y \mid} \sum_{y=Y_1}^{Y_{\mid Y \mid}} (\mathit{IRlbl}(y)) . \qquad\text{where} 
\label {Eq.MeanIR}
\end {equation} 
 
\begin {equation}\footnotesize
\mathit{IRlbl}(y)=\frac{\stackrel[{y^{\prime}=Y_1}]{Y_{\mid Y \mid}}{\operatorname{argmax}}\left(\sum_{i=1}^{|D|} h\left(y^{\prime}, Y_i\right)\right)}{\sum_{i=1}^{|D|} h\left(y, Y_i\right)} . \qquad\text{with} 
\label {Eq.IRlbl}
\end {equation}

\begin {equation}\footnotesize
\quad h\left(y, Y_i\right) = 
\begin{cases}
	1 &{y \in Y_i}\\
	0 &{y \notin Y_i}\\
\end{cases} .
\end {equation}

In addition to the way in which the IR is calculated, imbalanced MLDs manifest a number of distinctive characteristics, with the following two being of particular note:

\begin{itemize}
	\item \textbf{High imbalance levels}:
	In MLL, the degree of imbalance is typically much higher than in binary or multiclass scenarios. While a MeanIR greater than 10 is typically regarded as a high value in the context of traditional classification, it is not uncommon in the domain of MLL. Many multilabel datasets exhibit MeanIR values between 30 and 100, with some extreme cases exceeding 1000. This indicates that rare labels may occur only once for every several hundred instances of common labels, rendering it exceedingly challenging to train models that can effectively handle these low-frequency labels. The issue can be further aggravated when certain MLL techniques, such as binary relevance~\cite{Godbole}, are employed, as this results in the development of an independent binary model for each label in opposition to all the others.
	
	\item \textbf{Coupling of frequent and rare labels in MLL}:
	In MLDs, it is not uncommon for less frequent labels to appear in conjunction with majority labels within the same instances. This coupling, as quantified by the SCUMBLE (\ref{Eq.SCUMBLE}) metric~\cite{Charte:NeucomDifficultLabels}, introduces further complexities when attempting to address imbalance in MLL. In contrast to binary or multiclass classification, where resampling methods can be applied to individual classes, multilabel instances are associated with multiple labels. Consequently, the removal or generation of synthetic samples affects all labels associated with an instance, not merely the most common or rarest ones. The aforementioned coupling renders the reduction of imbalance ratios through traditional resampling techniques an arduous task. The augmentation of the number of samples containing rare labels frequently results in a proportional increase in common labels, thus leaving the overall imbalance ratio unaltered or even increased.
\end{itemize}

\begin {equation}\footnotesize
\mathit{SCUMBLE} = \frac{1}{{\mid D \mid}} \sum_{i=1}^{{\mid D \mid}} [1 - \frac{1}{\overline{\mathit{IRlbl_i}}} (\prod_{l=1}^{{\mid L \mid}} \mathit{IRlbl_{il}})^{\frac{1}{{\mid L \mid}}}] .
\label {Eq.SCUMBLE}
\end {equation}

As a result, the standard imbalance-handling techniques from binary and multiclass classification are frequently inapplicable to MLL due to the distinctive nature of multilabel data samples.

Traditionally, techniques such as data resampling, cost-sensitive learning, and algorithm-specific adaptations have been employed to address the issue of imbalanced classification~\cite{kotsiantis2006handling,mohammed2020machine}. As stated in Section~\ref{Sec.Introduction}, the former has the advantage of being model-independent, meaning it does not necessitate alterations to the model to enhance the learning process. Consequently, this approach will be the primary focus of our investigation.

\subsection{Oversampling Multilabel Data}

The resolution of imbalanced data through resampling methods represents a common preprocessing approach in the field of machine learning. Among these techniques, oversampling is typically preferred over undersampling due to its capacity to preserve and enhance the available information. Chawla et al.~\cite{Chawla2002} introduced SMOTE, a widely used oversampling method that generates synthetic minority instances, effectively increasing the representation of rare classes without discarding valuable data. In contrast, undersampling techniques, such as random undersampling (RUS) as discussed in~\cite{Liu2009}, entail the removal of instances belonging to the majority class, which can result in the loss of potentially crucial information.  A comprehensive review of various resampling methods, emphasizing the advantages of oversampling in maintaining data integrity, is provided by~\cite{he2009learning}. Moreover, hybrid approaches like SMOTEENN~\cite{Batista2004} integrate oversampling with selective undersampling, aiming to capitalize on the strengths of both techniques while mitigating their individual weaknesses. In general, the preference for oversampling stems from its capacity to augment the dataset without sacrificing existing samples, thereby preserving and potentially enhancing the model's ability to learn from all available information.

There are a number of different approaches that can be taken with regard to the implementation of a MOA. The following ones are particularly worthy of consideration:

\begin{itemize}
	\item \textbf{Random cloning:} The replication of instances that have been assigned the minority label represents a straightforward method for reducing imbalance levels. This approach is typically effective when working with binary and multiclass data, as each instance of the minority class can be cloned independently (it only has one label). In MOA, however, it is necessary to consider the possibility that a sample with a minority label may also be associated with one or more majority labels. Therefore, the cloning process will not necessarily result in a reduction of the imbalance level, and in certain instances, it may even lead to an increase. Some MOAs based on cloning are LPROS and MLROS~\cite{charte2015addressing}.
	
	\item \textbf{Synthetic generation:} This represents a more sophisticated methodology. In contrast to cloning existing instances, these MOAs generate new instances through the combination of feature values and labels, the computation of which is dependent on the specific method employed. The majority of these methods are based on the aforementioned SMOTE, although they include specific modifications to address the particular characteristics of MLDs. Two of them are MLSMOTE~\cite{mlsmote} and MLSOL~\cite{mlsol}.
	
	\item \textbf{Labelset decoupling:} The coupling of majority and minority labels in the same instances represents a significant challenge for MOAs. This approach entails the partitioning of the labelset associated with these instances, resulting in the creation of two distinct subsets: one comprising solely majority labels and the other exclusively with minority ones, thus creating two samples from one. This straightforward technique does not inherently guarantee an improvement in imbalance; however, it can be effectively integrated with other oversampling methodologies. REMEDIAL and its variants~\cite{remedial,rhwrsmt} exemplify this solution.
\end{itemize}

As previously stated, SMOTE-based MOAs are dependent on NN search of seed samples. The process of computing distances between all instances in order to determine which are the nearest is inherently time-consuming. The runtime is determined by the number of attributes and labels. The way in which each MOA produces the synthetic sample attributes and labels can also exert a certain impact.

\subsection{Diffusion Models}\label{Sec.DiffusionMOdels}

\begin{figure}[h!]
	\centering
	\includegraphics[width=\linewidth]{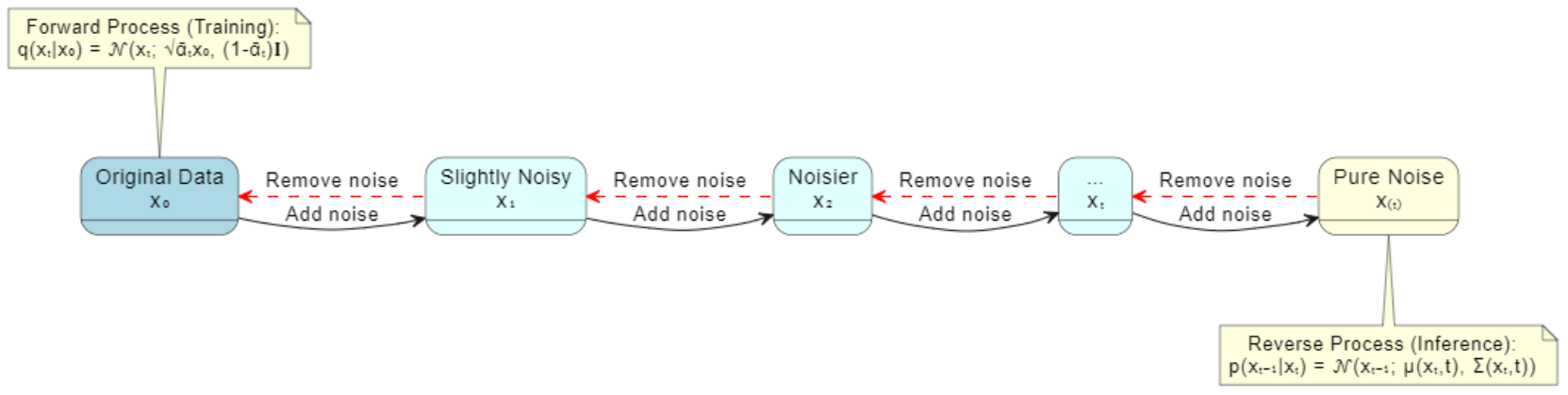}
	\caption{How a DDPM works: In training, noise is gradually added to the input while learning the reverse process. Inference produces a data sample from noise using the learned steps.}
	\label{Fig.DDPM}
\end{figure}

Over the past decade, a new field of study has emerged, termed generative AI. Many of these models, including variational autoencoders (VAEs)~\cite{kingma2013auto}, generative adversarial networks (GANs)~\cite{goodfellow2014generative}, normalizing flows (NMs)~\cite{rezende2015variational}, denoising diffusion probabilistic models (DDPMs)~\cite{sohl2015deep}, and more recently, transformers~\cite{vaswani2017attention}, have been primarily utilized for image and text generation. As is the case with most approaches, each of these has its own set of advantages and disadvantages. We choose DDPMs for a number of reasons:

\begin{itemize}
	\item They are able to produce higher quality data in comparison to VAEs without the issue of collapse during training.
	\item Their training process is more stable than that of GANs, where convergence is frequently challenging.
	\item They are not constrained by the limitations of NMs due to bijective transformations.
	\item They require fewer resources and time to be trained than transformers.
\end{itemize}

As illustrated in Fig.~\ref{Fig.DDPM}, a DDPM starts its training process by introducing Gaussian noise into the original data samples. This is an iterative process whereby a specific quantity of noise is introduced at each stage. Upon completion of the $T$ steps, the procedure yields pure noise that follows a known Gaussian distribution with mean value of $\mu_\theta$ and standard deviation of $\sigma_\theta$. A neural network is employed to determine the requisite parameters for the inverse process, which involves inverting each step until the original data distribution is achieved.

Equation (\ref{Eq.DDPM1}) is in charge of the forward process. It takes the sample $x$ in the time $t-1$ and adds a quantity of noise to obtain the sample in time $t$. The quantity is controlled by the term $\beta_t$, which sets the variance to add at each step. $\beta_t{I}$ denotes the covariance matrix of the Gaussian distribution. 

 \begin{equation}\footnotesize
 	q(x_t|x_{t-1}) = \mathcal{N}(x_t; \sqrt{1-\beta_t}x_{t-1}, \beta_t I) .
 	\label{Eq.DDPM1}
 \end{equation}
 
 The reverse process (\ref{Eq.DDPM2}) determines the conditional probability distribution which allows obtaining $x_{t-1}$ from $x_t$ through a denoising mechanism. $\mu$ (mean) and $\Sigma$ (covariance matrix) are predicted by an artificial neural network (ANN), allowing the sampling of the $\mathcal{N}$ distribution at each step.
 
 \begin{equation}\footnotesize
 	p_\theta(x_{t-1}|x_t) = \mathcal{N}(x_{t-1}; \mu_\theta(x_t, t), \Sigma_\theta(x_t, t)) .
 	\label{Eq.DDPM2}
 \end{equation}

The ANN learning is guided by the (\ref{Eq.Loss}) loss function. The Kullback-Leibler divergence ($D_{KL}$) is used to obtain the difference between the true reverse process ($q(x_{t-1} | x_t)$) and the one being learned by the ANN ($p_\theta(x_{t-1} | x_t)$). The average expected value for this difference allows to compute the committed error.

\begin{equation}\footnotesize
	L = \mathbb{E}_q \left[ \sum_{t=1}^T D_{KL}(q(x_{t-1} | x_t) \parallel p_\theta(x_{t-1} | x_t)) - \log p_\theta(x_0) \right] .
	\label{Eq.Loss}
\end{equation}

\section{MLDM: A Diffusion-based MOA Model}\label{Sec.MLDM}

Once the background has been established, this section is devoted to describing the proposed MLDM algorithm. Initially, a high-level overview of its functioning is provided. Subsequently, the constituent elements of its internal workings are elucidated. Finally, the computational complexity of MLDM is compared with that of other MOAs.

\subsection{High-level overview}

%%{init: {'theme': 'base'}}%%
%flowchart 
%V["Multilabel dataset"] --> W["Extract minority samples"]
%W --> X["Training"]
%X -->|Model| Y["MLDM"]
%Y ==> Z["Synthetic samples"]
%B["Gaussian noise"] ==>|Sampling| Y
%
%style V fill:#f9f,stroke:#333,stroke-width:2px
%style B fill:#f9f,stroke:#333,stroke-width:2px
%style Y fill:#aaa,stroke:#ccc,stroke-width:4px
%style Z fill:#bbf,stroke:#f66,stroke-width:2px

\begin{figure}[h!]\centering
	\includegraphics[width=.7\linewidth]{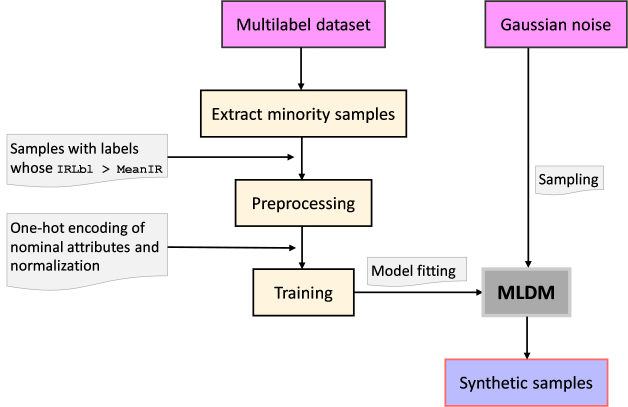}
	\caption{The process begins with any MLD, from which minority samples are extracted and utilized to train the model. Subsequently, the model is capable of generating synthetic samples, beginning with Gaussian noise.}
	\label{Fig.MLDMOverview}
\end{figure}

The main steps in training and using MLDM are illustrated in Fig.~\ref{Fig.MLDMOverview}. The left branch of the diagram is in charge of generating the model. To do so, samples with at least one minority label are obtained. These are preprocessed and then used in the training algorithm to fit the generative model. The right side shows how synthetic samples are created from pure noise.

Note that only one model per dataset is trained. This decision was made after analyzing three possible approaches:

\begin{enumerate}
	\item \textbf{Train a global model} to learn the entire MLD distribution, so that it is able to generate synthetic instances linked to all labels. Then, add only those with minority labels to the dataset. This approach was discarded since most of the instances produced would contain majority labels, making it difficult to reduce the imbalance level.
	
	\item \textbf{Train individual models}, one for each minority label, and then use each of these models to create new instances with only those labels. The main drawback of this technique is that in many MLDs there are only a few instances for some labels. These would not be enough to train a good model.
	
	\item \textbf{Train a model specialized in minority labels}, a strategy that is a mixture of the previous ones. By collecting all instances with minority labels, a larger subset of samples can be obtained and thus a better model can be trained. By avoiding the cases where only majority labels make up the labelset, such as in method 1, it is guaranteed that the model will focus on generating patterns associated with the labels of interest.
\end{enumerate}

\subsection{Internal workings}

Having considered the foundations of MLDM as presented in the overview, the following sections will elaborate on the aspects that are of particular significance within its operational context.

\subsubsection{Minority samples selection}

The method begins with the MLD provided as input and proceeds to select a subset (\textit{trainSubset}) of samples for training the model. The decision of whether a sample is included in the subset is based on a straightforward criterion: it must contain one or more labels $l$ whose $\textit{IRLbl(l)} >  \textit{MeanIR}$, i.e. the frequency of label $l$ in the MLD is lower than the average frequency.

\subsubsection{Attributes encoding}

Once the \textit{trainSubset} has been obtained, a one-hot encoding~\cite{potdar2017comparative} is applied to all nominal attributes, including the labels. Similarly, numeric attributes undergo a normalization process. Specifically, a quantile normalization process, as described in~\cite{peterson2020ordered}, is applied. This approach allows using of both discrete and continuous variables as numeric inputs for the training of the diffusion model. 

\subsubsection{Model training}

The training of MLDM follows the steps described in section~\ref{Sec.DiffusionMOdels}, common to any standard diffusion model. However, there are two major differences from other proposals. The first is that we use all attributes, including labels, to train our model. This approach differs from other DDPMs used with tabular data, where the class is excluded from training, so that the generative model produces only the set of attributes, while the label is determined afterwards. Once trained, MLDM generates complete instances, including all attributes and labels. The second difference lies in the distribution used in the process, where the usual Gaussian has been changed to a multinomial distribution~\cite{hoogeboom2021argmax} aiming to model the disparate distributions followed by numeric and nominal attributes. This latter decision was inspired by the TabDDPM model~\cite{kotelnikov2023tabddpm}, but unlike it, MLDM does not separate numeric attributes from nominal ones, but transforms them through one-hot encoding as explained above.

\subsubsection{Synthetic samples generation}

Once the model is fitted, it is used to generate the synthetic samples needed to balance the MLD. Starting with Gaussian noise, new samples are produced until the percentage specified by the $D$ parameter is reached. The attributes of each synthetic instance are adjusted through an inverse normalization process so that they return to their original range of values. The one-hot encoded labels are also processed to create the corresponding labelset.

\subsection{Computational Complexity}

The two primary objectives of the MLDM design were to generate high-quality synthetic samples and to be efficient. While the actual running times will be evaluated at a later stage, this section presents a concise comparison of the computational complexity of MLDM and other MOAs.

First of all, it has to be considered whether the resampling methods to be analyzed work in a single stage, only doing sample generation, or in two stages, fitting a model and then doing sample generation.  Simple approaches, such as LPROS and MLROS, are very fast because they simply clone existing instances after verifying that they contain some of the minority labels. The computational complexity of these methods is linear to the number of samples to generate. Algorithms based on SMOTE, such as MLSOL and MLSMOTE, do not build a model, but work in two steps: first, the subset of samples with minority labels is extracted; then, new samples are generated by NN search. If the dataset has $n$ samples and $m$ of them are on the minority label subset, with $t$ attributes, the first step has complexity $O(n)$, while the second step has complexity $O(m^2t)$ if distances between all samples are needed. The latter factor must be multiplied by the number of new samples to be generated. The number of labels $l$ in the MLD could also have some impact on complexity if it is large. Some MLDs have thousands of labels, in fact there are methods to reduce the label dimensionality such as~\cite{LI-MLC}, and $l$ could outweigh the importance of the other factors. The number of NNs to be obtained, $k$, is always a small value, so it should not have a significant impact on the runtime.

MLDM, as the name implies, builds a model before it starts to create synthetic samples, i.e. it works in two steps. The first trains the diffusion model and is run only once. These models create an affinity matrix of pairwise similarities between samples, with a size of $m\times m$ ---we train MLDM only with the subset of minority samples and, in general, $m<<n$--- and a time complexity of $O(m^2t)$. This matrix is the basis for iteratively updating the internal representations of the model by $c$ steps, where $c$ is a multiplicative factor of the previous complexity. The process of generating a new sample is several orders of magnitude less complex than building the model, and linear in the number of instances generated. This process does not require searching for NNs, as in SMOTE-based proposals, nor computing the $l$ labels for the new sample, since all the information is obtained at once.

\section{Experimentation and Results}\label{Sec.Experiments}

This section first describes the experimental setup used to evaluate the performance of MLDM, including the datasets, classifiers and metrics used, as well as the oversampling methods compared with MLDM.

\subsection{Experimental setup}

MLL experiments are time consuming and resource intensive. Most classifiers rely on data transformations and ensembles of models, which need large amounts of memory and run time. The resampling methods themselves are also costly, as the imbalance level of each label has to be computed, and many of them use NN search for each instance. On the other hand, a robust experiment should use as many datasets, methods and classifiers as possible. Looking for a balance between these two perspectives, our experimental setup is a combination of 8 MLDs, 6 resampling methods and 5 different classifiers. Each of these 240 configurations was run with 5-fcv, resulting in a total of 1\,200 runs.

\subsubsection{Datasets}
	
The eight datasets whose properties are shown in Table~\ref{Tbl.MLDMetrics} were downloaded from the Cometa MLD repository~\cite{cometaml}. Specifically, we used the 5-fcv partitions provided in this repository, so that anyone interested can use exactly the same data partitions to reproduce our experiments. Since the goal is to test how MLDM deals with the imbalance problem, most of the MLDs chosen have high MeanIR and SCUMBLE. The exceptions are emotions and scene, two classic MLDs (they can be found in almost every MLL paper), which were included as a baseline performance control.

\begin{table}[h!]
	\centering\footnotesize\setlength{\tabcolsep}{2pt}
		\begin{tabular}{lrrrrrrrr}
			\toprule
			\textbf{Name} & \textbf{\#inst.} & \textbf{\#attr.} & \textbf{\#labels} & \textbf{Card} & \textbf{Dens} & \textbf{MeanIR} & \textbf{Scumble} & \textbf{Ref.} \\
			\midrule
			{cal500} & 502 & 68 & 174 & 26.0438 & 0.1497 & 20.5778 & 0.3372  & \cite{cal500} \\
			{corel5k} & 5\,000 & 499 & 374 & 3.5220 & 0.0094 & 189.5676 & 0.3941  & \cite{corel5k} \\
			{emotions} & 593 & 72 & 6 & 1.8685 & 0.3114 & 1.4781 & 0.0110 &  \cite{emotions} \\
			{genbase} & 662 & 1\,186 & 27 & 1.2523 & 0.0464 & 37.3146 & 0.0288 &  \cite{genbase} \\
			{medical} & 978 & 1\,449 & 45 & 1.2454 & 0.0277 & 89.5014 & 0.0471 &  \cite{medical} \\
			{scene} & 2\,407 & 294 & 6 & 1.0740 & 0.1790 & 1.2538 & 0.0003 &  \cite{scene} \\
			{chess} & 1\,675 & 585 & 227 & 2.4113 & 0.0106 & 85.7898 & 0.2625 &  \cite{stackexchess} \\
			{yeast} & 2\,417 & 103 & 14 & 4.2371 & 0.3026 & 7.1968 & 0.1044 &  \cite{yeast} \\
			\bottomrule
		\end{tabular}
	\caption{Main characterization metrics for the MLDs used in the experiments (source:~\cite{cometaml}). The columns provide the name of the MLD, number of instances, attributes and labels, label cardinality and density, its mean IR, the level of label coupling (SCUMBLE), and the corresponding bibliographic reference.}
	\label{Tbl.MLDMetrics}
\end{table}	
	
\begin{figure}[h!]
	\centering
	\includegraphics[width=.7\linewidth]{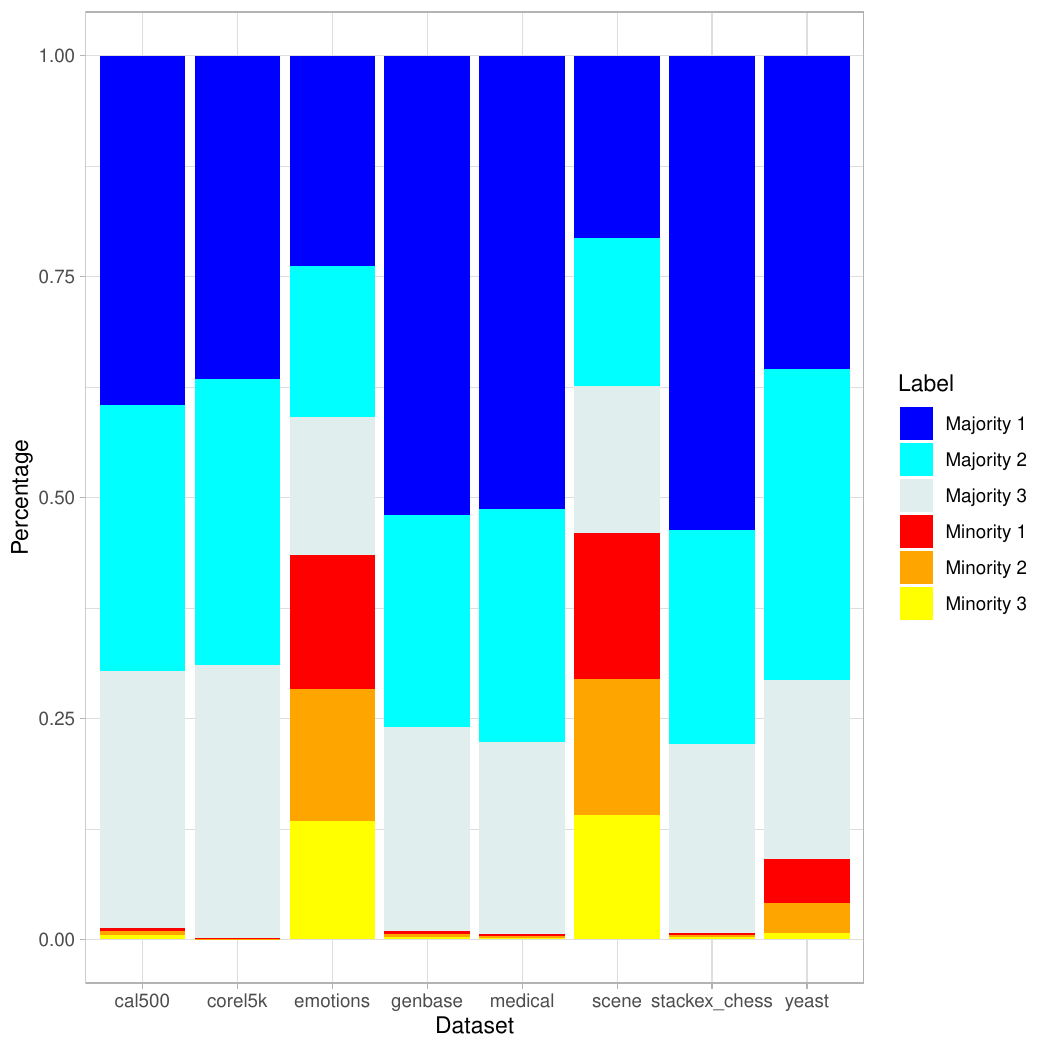}
	\caption{Frequency of the three most common labels (blue shades) and the three rarest labels (red shades) in each MLD. In most cases, the total number of instances with minority labels is almost negligible compared to the majority ones.}
	\label{Fig.Frequencies}
\end{figure}

Sometimes raw numbers do not give a clear idea of the differences between label frequencies that a particular MeanIR implies. Fig.~\ref{Fig.Frequencies} shows the relative frequency of six labels in each MLD, the three most common and the three rarest. The difficulty of learning to classify minority labels with such large differences can be seen from the darkest shades at the bottom, which are almost imperceptible for some datasets.
	
\subsubsection{Resampling methods}
	
	Fall into two main groups, oversampling and undersampling. The former creates new instances with minority labels, while the latter removes data points with majority labels. Both tend to produce more balanced datasets, but it is generally accepted that undersampling can remove potentially useful samples~\cite{he2009learning,japkowicz2002class,lopez2013insight}, leading to worse results. For this reason, we propose an oversampling method, and the following experiments use only algorithms that follow this approach. Table~\ref{Tbl.Algorithms} lists the algorithms to be compared, indicating their type and some common parameters.
		
	\begin{table}[h!]
		\centering\footnotesize
		\begin{tabular}{llccc}
			\toprule
			\textbf{Name} & \textbf{Type} & \textbf{P (\%)} & \textbf{\textit{k}} & \textbf{Ref.}\\
			\midrule
			{LPROS} & Random labelset clonning & 25 & - & \cite{charte2015addressing}\\
			{MLROS} & Random label clonning & 25 & - & \cite{charte2015addressing}\\
			{MLSMOTE} & Synthetic (nearest neighbors) & - & 3 & \cite{mlsmote}\\
			{MLSOL} & Synthetic (nearest neighbors) & 25 & 3 & \cite{mlsol}\\
			{REMEDIAL} & Labelset splitting & - & - & \cite{remedial}\\
			{\textbf{MLDM}} & Synthetic (diffusion model) & 25 & - & - \\
			\bottomrule
		\end{tabular}
		\caption{Resampling algorithms compared in the experiments. \textit{P} denotes the percentage of synthetic new samples to create with respect to the original size of the MLD, while \textit{k} is the number of instances for algorithms that that use NN.}
		\label{Tbl.Algorithms}
	\end{table}
	
	These algorithms have already been described above. They were chosen to provide a wide variety of strategies for generating new samples associated with minority labels, including instance cloning, synthetic generation, sample labelset decoupling, etc. 
	
	%The implementation of these algorithms used in our experiments is provided by mldr.resampling~\cite{mldr.resampling}. This is an R package that is publicly available in the official R repository CRAN\footnote{\url{https://cran.r-project.org/web/packages/mldr.resampling/}}, so anyone can reproduce the same steps. MLDM is not yet included in this package, but its source code is available on our research group's Github page\footnote{\url{https://github.com/madr0008/mldm}}.
	
\subsubsection{Classifiers}

	To assess the effectiveness of each resampling method, once the training partitions have been processed, they are given as input to a set of classifiers. Since not all methods are equally affected by imbalance, this set includes a variety of approaches to MLC, as shown in Table~\ref{Tbl.Classifiers}.

	\begin{table}[h!]\footnotesize
		\centering
		\begin{tabular}{llc}
			\toprule
			\textbf{Name} & \textbf{Type} & \textbf{Ref.}\\
			\midrule
			BR-J48 & Binarization with C4.5 & \cite{br.lp}\\
			LP-J48 & Label powerset with C4.5 & \cite{br.lp}\\
			BPMLL & Adapted neural network & \cite{bpmll} \\
			MLkNN & Adapted nearest neighbor & \cite{mlknn} \\
			HOMER & Hierarchical ensemble & \cite{homer} \\
			\bottomrule
		\end{tabular}
		\caption{Set of classifiers. Each one is trained with the corresponding partitions, both in their original state and after being processed with each resampling method.}
		\label{Tbl.Classifiers}
	\end{table}
	
	The binarization approach (BR-J48) is highly impacted by imbalance levels, since a separate binary classifier is trained for each label. Therefore, some of them are trained with only a handful of positive samples. An effective oversampling method should improve the behavior of these minority classifiers by training them with more positive data points.
	
	With LP-J48 only one classifier is trained, using each different labelset as a class. The labelsets containing minority labels will be rare, especially if the MLD has a large SCUMBLE, as the instances with minority labels will mostly also be associated with majority labels. 
	
	Neural networks such as BPMLL are sensitive to imbalance due to their training process. The most common patterns, i.e. majority labels, have more influence in gradients than those rarely seen, i.e. minority labels. Nevertheless, BPMLL should have a more global view of the data than the former approaches, so the impact of imbalance and resampling in their results might not be as important as with BR or LP.
	
	As the BR approach, MLkNN is highly impacted by imbalance level. The MLkNN algorithm searches for neighbors of the given instance to be classified. Since most of the space is occupied by samples with majority labels, the probability of misclassifying the minority ones is very high. 
	
	Ensembles tend to be resilient to imbalance and are often proposed as an effective approach for dealing with this type of data. HOMER is a model built as an ensemble of classifiers trained on subsets of the original labelsets, so it should be less affected by the imbalance levels.

\subsubsection{Performance metrics}

	A few dozen performance metrics are defined in the MLL literature. Each provides a different perspective on the outcomes being analyzed~\cite{CharteMLC}. There are ranking-based and bipartition-based metrics. Some are calculated by instance (sample-based) and then averaged, while others are computed by label (micro- and macro-averaged). A common problem in MLL studies is how to interpret the results when many performance metrics are used to evaluate the classifiers and they give conflicting views.
	
	\begin{table}[h!]
		\centering\footnotesize
		\begin{tabular}{lllc}
			\toprule
			\textbf{Name} & \textbf{Category} & \textbf{Averaging} & \textbf{Eq.}\\
			\midrule
			Hamming loss (HL) & Sample based & Global & \ref{HL}\\
			F-measure (F1) & Sample based & Global & \ref{F1}\\
			Macro-F1 & Label based & Macro & \ref{MacroM}\\
			Micro-F1 & Label based & Micro & \ref{MicroM}\\
			One error (OE) & Ranking based & Global & \ref{OneError}\\
			\bottomrule
		\end{tabular}
		\caption{Performance metrics used in the experiments and their characteristics.}
		\label{Tbl-Metrics}
	\end{table}
	
	In the present experiments, the interest is not to compare the classifiers, but to see how they improve depending on the resampling applied to the MLDs. We decided to include the metrics summarized in Table~\ref{Tbl-Metrics}: two sample-based metrics (Hamming loss and F-measure), a label-based metric with different averaging (Macro-F1 and Micro-1), and a ranking-based metric (One error). By choosing F-measure as the reference metric, we are assessing a balance (the harmonic mean) between precision and recall from three different perspectives, so it would be as if there were six measurements instead of three. However, having fewer values makes the results easier to analyze.
	
	\begin{equation}\footnotesize
		\textit{HL} = \frac{1}{|D|} \displaystyle\sum\limits_{i=1}^{|D|} \frac{|Y_i \Delta Z_i|}{|L|} . \label{HL}
	\end{equation}
	
	\begin{equation}\footnotesize
	Precision = \frac{1}{\mid D \mid} \displaystyle\sum\limits_{i=1}^{\mid D \mid} \frac{\mid Y_i \cap Z_i \mid }{\mid Z_i \mid } . \qquad
%	\end{equation}
%	
%	\begin{equation}\footnotesize
		Recall = \frac{1}{\mid D \mid } \displaystyle\sum\limits_{i=1}^{\mid D \mid } \frac{\mid Y_i \cap Z_i \mid}{\mid Y_i \mid} .
	\end{equation}
	
	\begin{equation}\footnotesize
		\textit{F1} = 2 \cdot \frac{Precision \cdot Recall}{Precision + Recall} \label{F1} .
	\end{equation}
	
	\begin{equation}\footnotesize\label{MacroM}
		\textit{Macro-F1}=\frac{1}{|L|} \sum\limits_{i=1}^{|L|}\textit{F1}(\textit{TP}_i,\textit{FP}_i,\textit{TN}_i,\textit{FN}_i) .
	\end{equation}
	
	\begin{equation}\footnotesize\label{MicroM}
		\textit{Micro-F1}=\textit{F1}(\sum\limits_{i=1}^{|L|}\textit{TP}_i,\sum\limits_{i=1}^{|L|}\textit{FP}_i,\sum\limits_{i=1}^{|L|}\textit{TN}_i,\sum\limits_{i=1}^{|L|}\textit{FN}_i) .
	\end{equation}

	\newcommand{\opA}{\mathop{\vphantom{\sum}\mathchoice
			{\vcenter{\hbox{\large argmax}}}
			{\vcenter{\hbox{\large argmax}}}{\mathrm{argmax}}{\mathrm{argmax}}}\displaylimits}
	\begin{equation}\footnotesize
		\textit{OE} = \frac{1}{n} \displaystyle\sum\limits_{i=1}^{n} \llbracket [\opA\limits_{y \in Z_i} \langle rk(x_i, y) \rangle \notin Y_i] \rrbracket . \label{OneError}
	\end{equation}
	
	In these equations the following notations are used: $Y_i$ denotes the real labelset of an instance, whereas $Z_i$ is the predicted one.

\subsubsection{Hardware and software}

	Both resampling algorithms and classifiers were run in the same hardware/software configuration. The machine has an Intel Xeon Silver 4210R CPU with 20 cores (40 threads), 256 GB of main memory and an NVidia A100 GPU with 80 GB of memory and 6\,912 CUDA cores. The operating system is Debian GNU/Linux with corresponding NVidia drivers and CUDA libraries.
	
	The mldr.resampling R package~\cite{mldr.resampling} was used to run the MOAs. This package provides reference implementations for many of the existing MLL resampling methods. The proposed MLDM algorithm is not yet included in this package, but we plan to add it in the future. In the meantime, the MLDM code is available as a Github repository\footnote{\url{https://github.com/simidat/mldm}}.
	
	The MLC methods are all available in the Mulan Java package~\cite{MULAN}. It is considered the reference software for multilabel classification. We fed each algorithm with the specific training and test partitions obtained from Cometa, and Mulan returned the requested set of performance metrics.
	
\subsection{Results}

After running all the classifiers on the datasets before and after applying the different resampling algorithms, three sets of results were collected:
\begin{itemize}
	\item The new imbalance levels of the datasets after preprocessing them with each oversampling method.
	\item The performance measures given by the classifiers.
	\item The runtime for each oversampling method.
\end{itemize}

\subsubsection{Changes in imbalance levels}

\begin{figure}[h!]
	\centering
	\includegraphics[width=.7\linewidth]{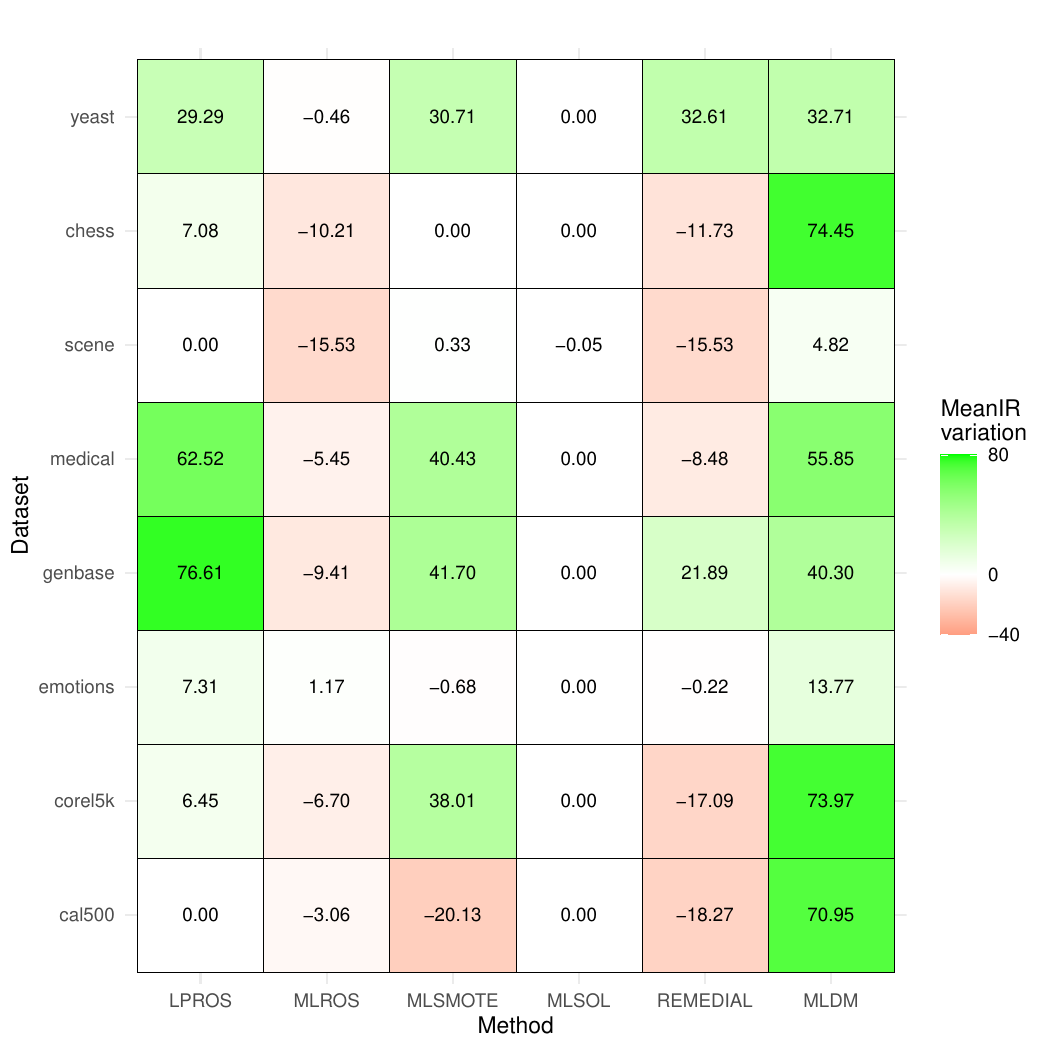}
	\caption{Heatmap of changes in \textit{MeanIR} for each dataset and resampling method. Positive values denote an improvement, while negative ones indicate a deterioration (i.e. the level of imbalance has increased).}
	\label{Fig.MeanIR}
\end{figure}

The aim of a MOA is to reduce the amount of imbalance present in the data. It is therefore of interest to analyze how the \textit{MeanIR} changes after it has been applied. To facilitate this analysis, we calculated the percentage improvement (or deterioration) of the \textit{MeanIR} metric with respect to its original value for each dataset/resampling method combination. The resulting values are presented in a heatmap (see Fig.~\ref{Fig.MeanIR}), with precise values within each cell.

\subsubsection{Classification improvements}

Once the datasets had been processed with each of the MOAs, the classifiers were run and the above metrics were obtained. The set of 1\,400 values, properly grouped, are provided in tables as additional material.%in Appendix \ref{Sec.Results}.

\begin{figure*}[h!]
  \renewcommand{\arraystretch}{0}\centering
  \setlength{\tabcolsep}{4pt}\footnotesize
  \begin{tabular}{p{1.7cm}p{2cm}p{2cm}p{2cm}p{2cm}p{2cm}p{2cm}}
  	\textbf{Method $\rightarrow$ Metric $\downarrow$} & 
  	   {\hspace{.7cm}\textbf{LPROS}} & 
  	   {\hspace{.8cm}\textbf{MLDM}} & 
  	   {\hspace{.7cm}\textbf{MLROS}} & 
  	   {\hspace{.4cm}\textbf{MLSMOTE}} & 
  	   {\hspace{.7cm}\textbf{MLSOL}} & 
  	   {\hspace{.3cm}\textbf{REMEDIAL}} \\
  	 \begin{tabular}{r} \textbf{HL} \end{tabular} & 
  	 \begin{tabular}{c} {\includegraphics[width=13.5cm]{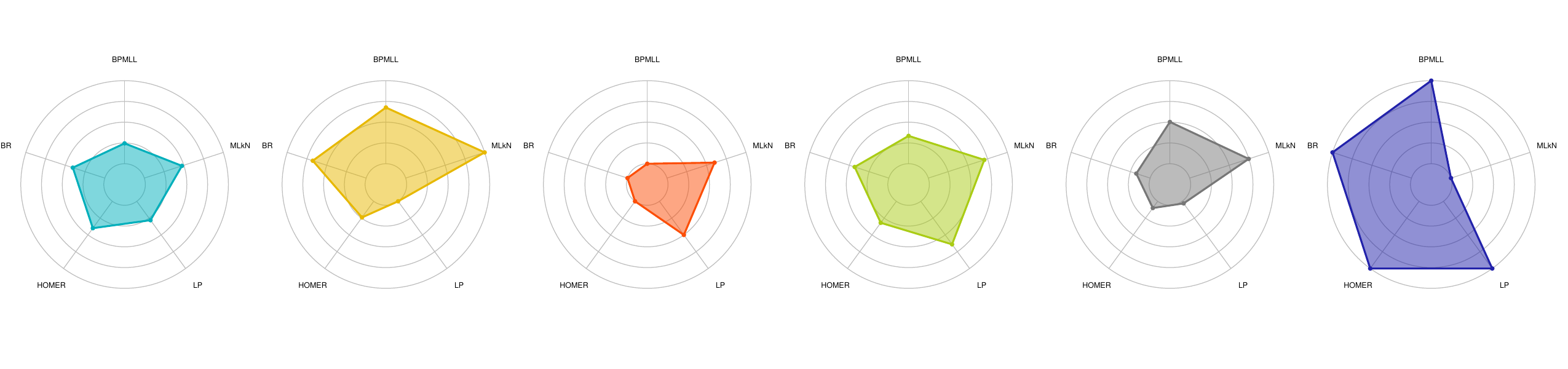}} \end{tabular} \\
  	 \begin{tabular}{r} \textbf{F1} \end{tabular} & 
  	 \begin{tabular}{c} {\includegraphics[width=13.5cm]{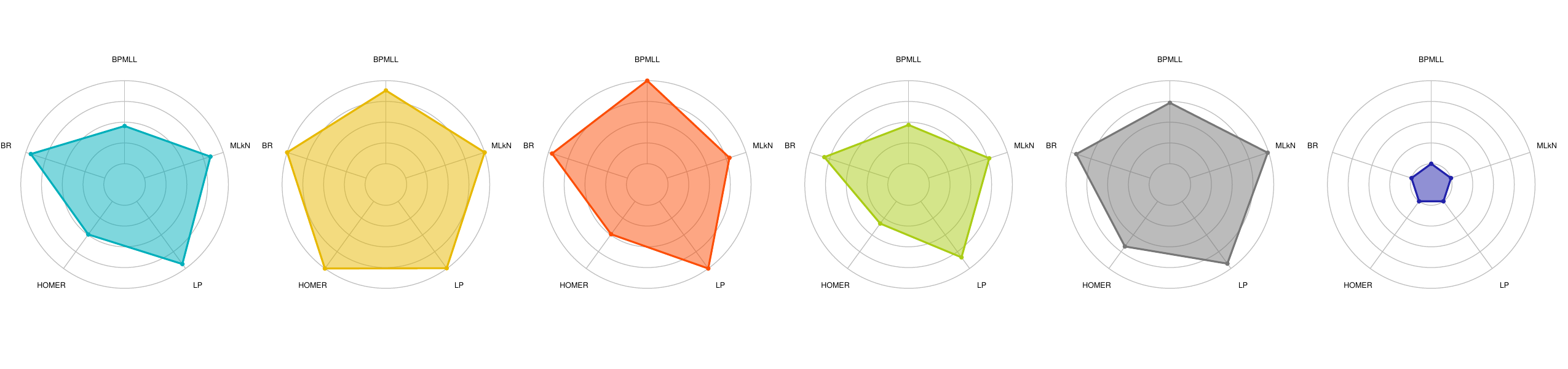}} \end{tabular} \\
  	 \begin{tabular}{r} \textbf{Macro-F1} \end{tabular} & 
  	 \begin{tabular}{c} {\includegraphics[width=13.5cm]{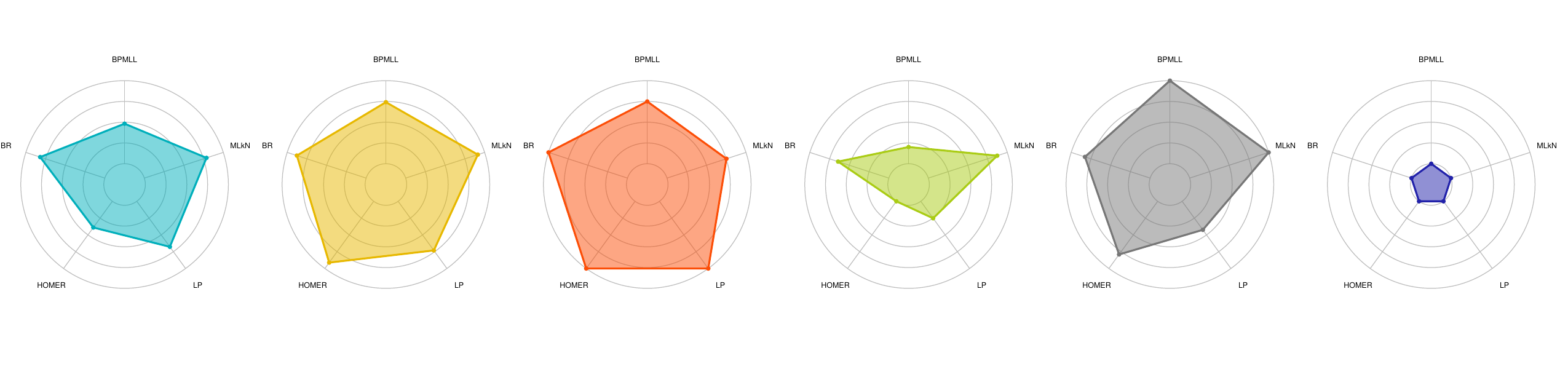}} \end{tabular} \\
  	 \begin{tabular}{r} \textbf{Micro-F1} \end{tabular} & 
  	 \begin{tabular}{c} {\includegraphics[width=13.5cm]{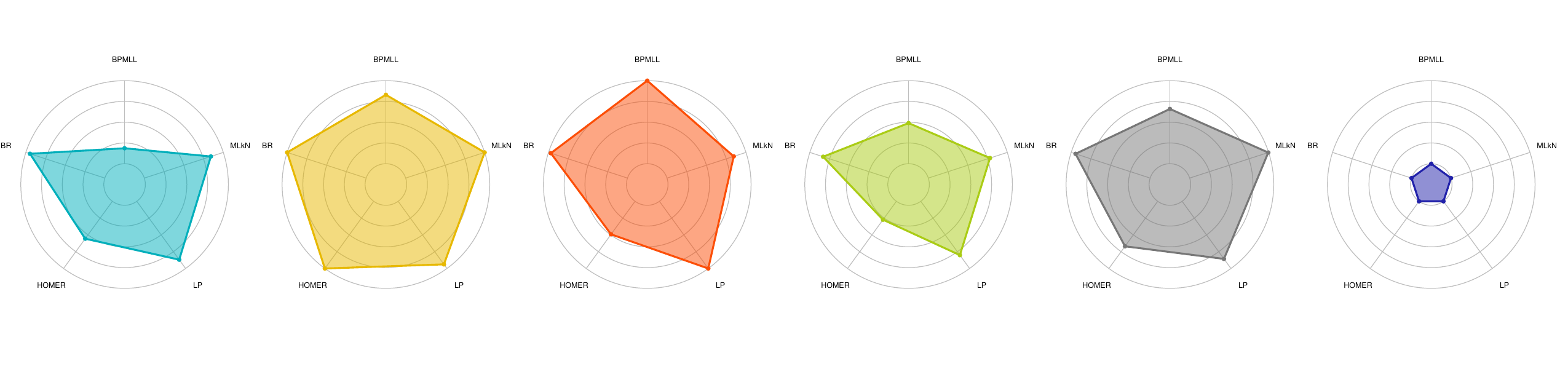}} \end{tabular}\\
  	 \begin{tabular}{r} \textbf{OE} \end{tabular} & 
  	 \begin{tabular}{c} {\includegraphics[width=13.5cm]{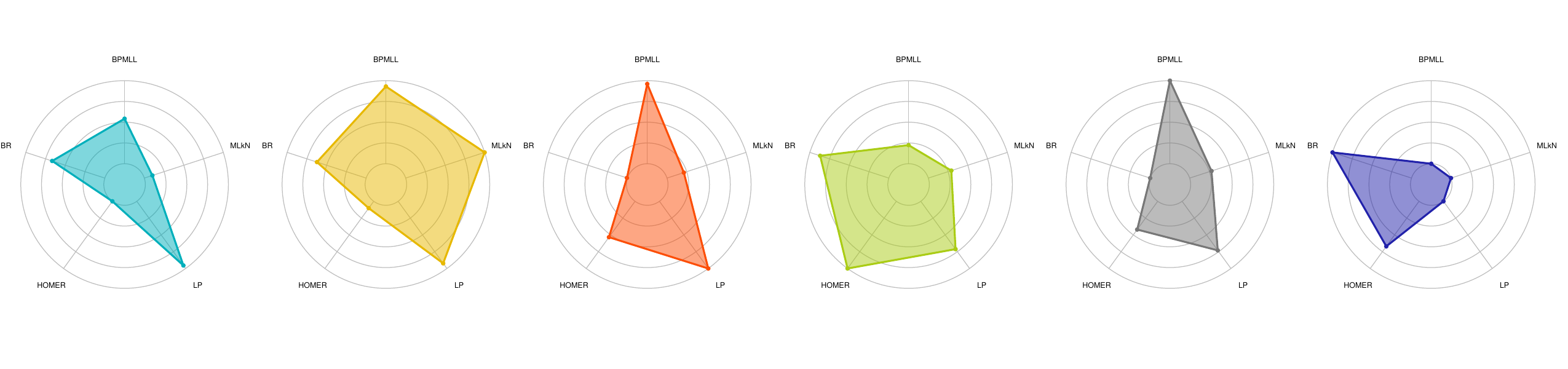}} \end{tabular} \\
  \end{tabular}	
  \caption{Radar plots showing classification performance by metric (row), resampling method (column) and classifier (vertex). The larger the area, the better the performance.}
  \label{Fig.RadarPlot}
\end{figure*}

To make it easier to analyze these results, they are presented as radar plots in Fig.~\ref{Fig.RadarPlot}, with the metrics as rows and the resampling methods as columns. For each plot, the classifiers appear as vertices. The larger the colored area, the better the performance. 

\begin{figure*}[h!]
	\centering
	\begin{subfigure}{0.195\textwidth}
		\includegraphics[width=\linewidth]{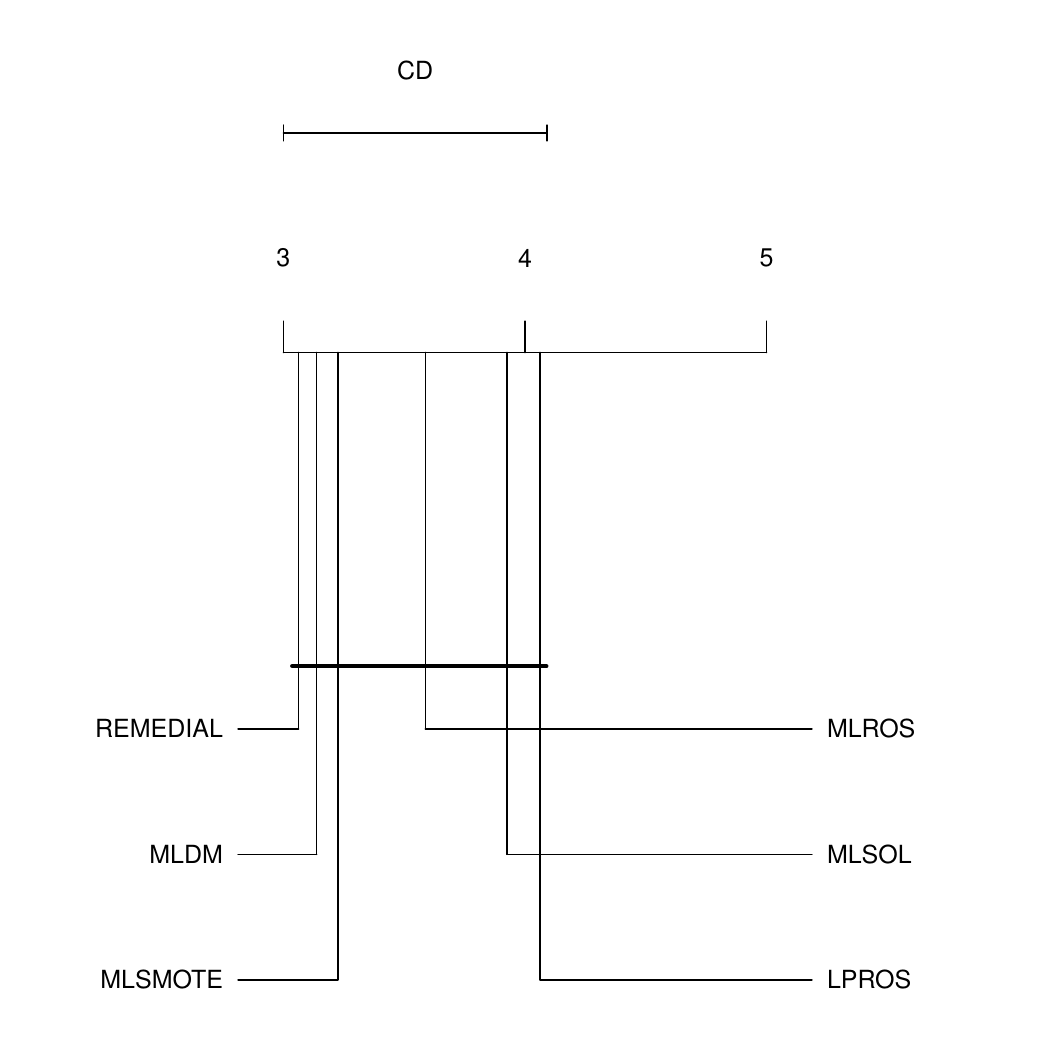}
		\caption{Hamming loss}
	\end{subfigure}
	\begin{subfigure}{0.195\textwidth}
	\includegraphics[width=\linewidth]{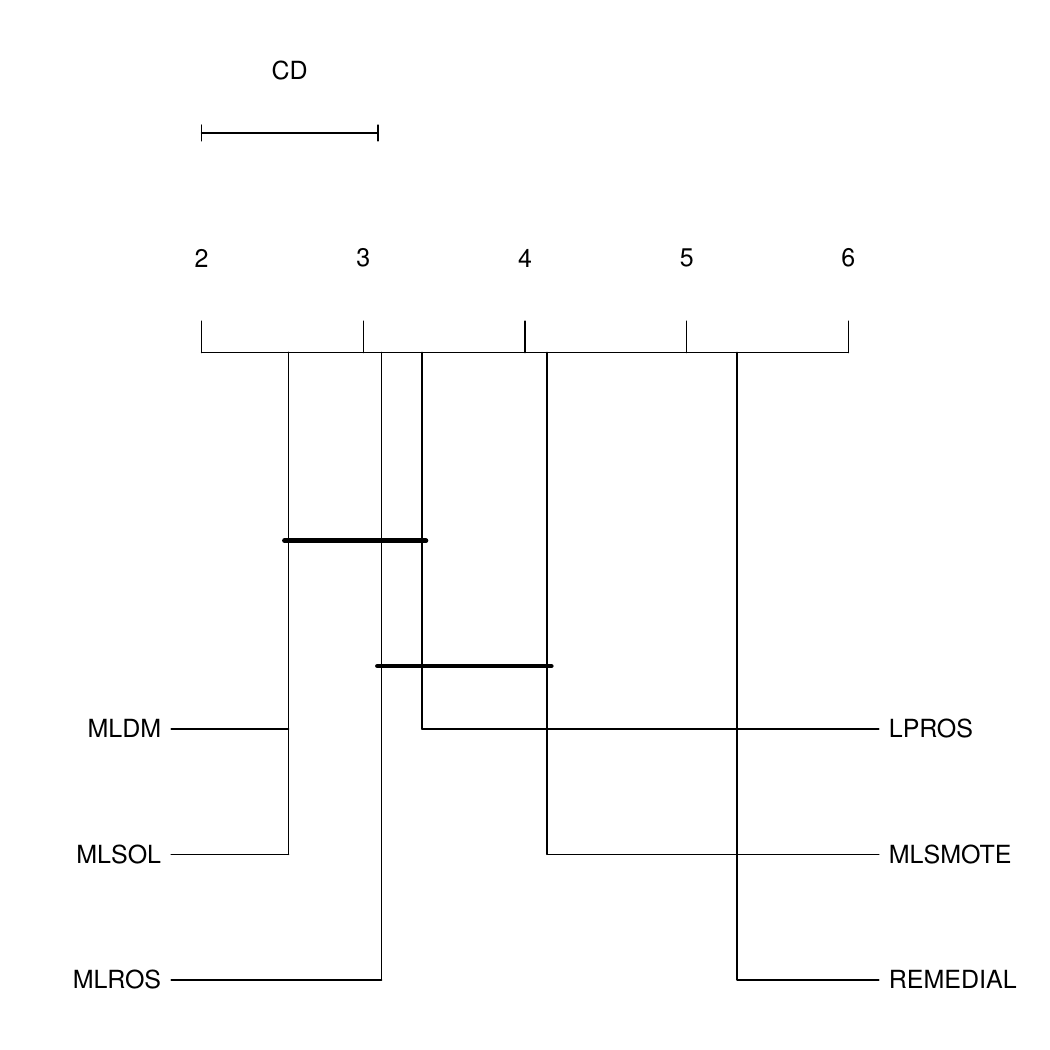}
		\caption{F1}
	\end{subfigure}
	\begin{subfigure}{0.195\textwidth}
	\includegraphics[width=\linewidth]{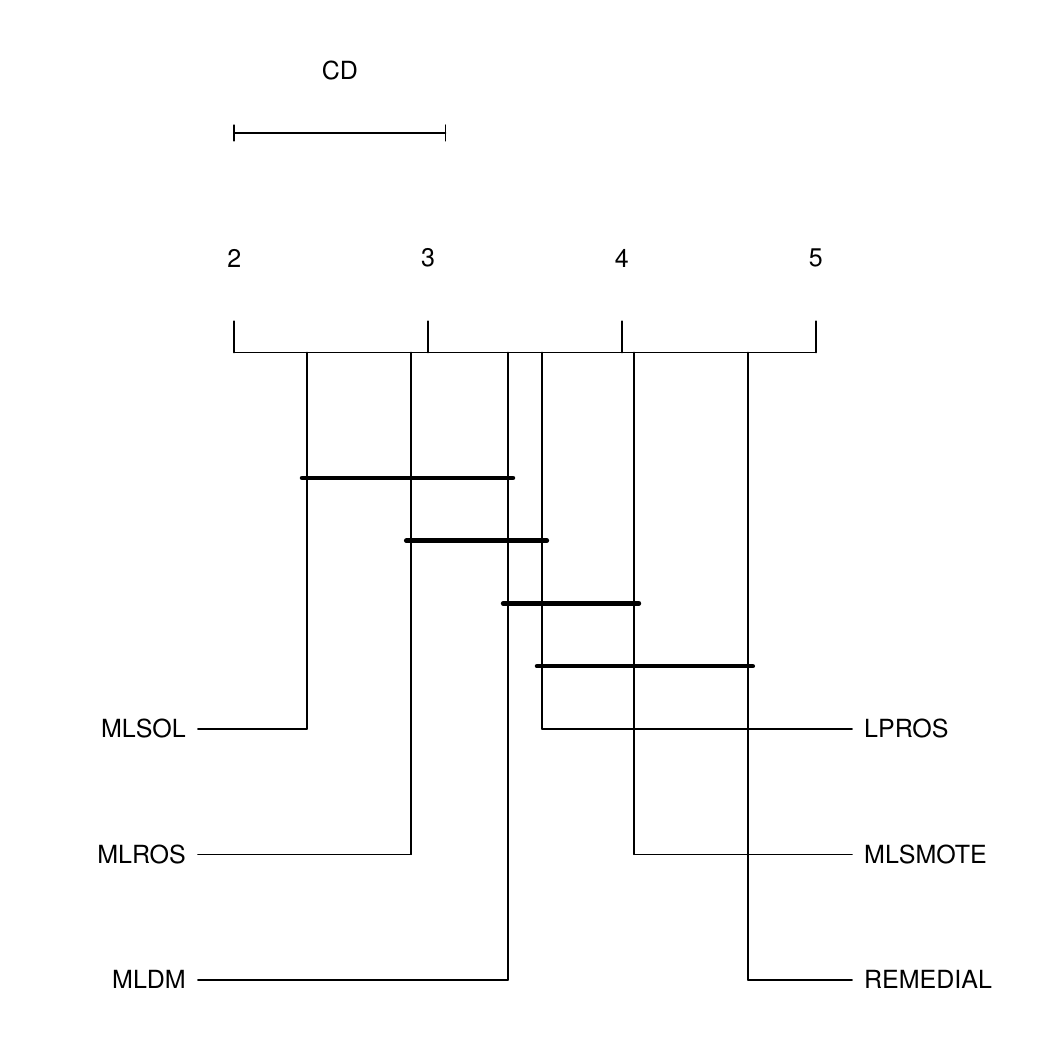}
		\caption{Macro F1}
	\end{subfigure}
	\begin{subfigure}{0.195\textwidth}
	\includegraphics[width=\linewidth]{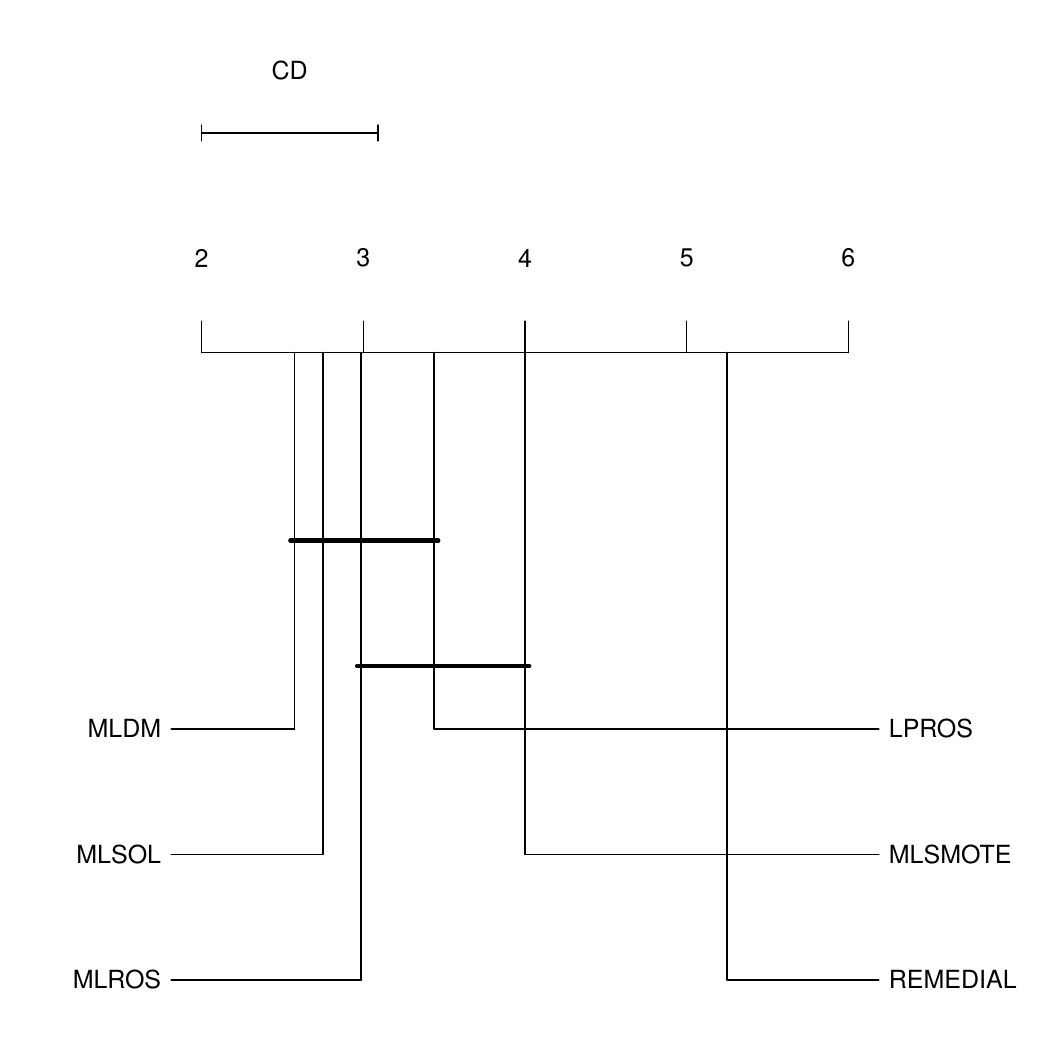}
		\caption{Micro F1}
	\end{subfigure}
	\begin{subfigure}{0.195\textwidth}
	\includegraphics[width=\linewidth]{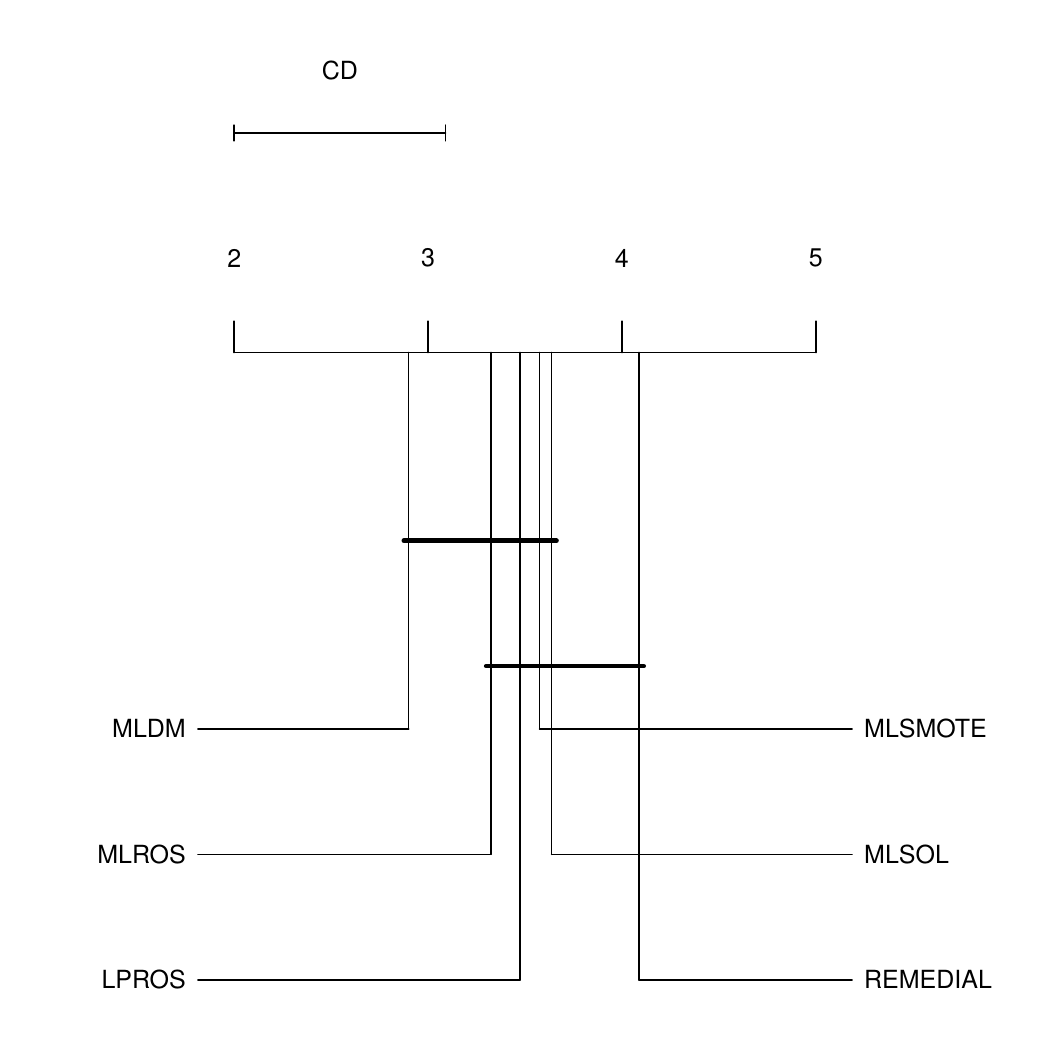}
		\caption{One error}
	\end{subfigure}
	\caption{Critical distance plots for each performance metric. As can be observed, both MLDM and MLSOL tend to appear as the best ranked algorithms. However, there are no statistical differences between them.}
	\label{Fig.CD}
\end{figure*}

To assess the statistical significance of differences among these results, the Nemenyi statistical test~\cite{demvsar2006statistical} was applied. Significant differences were found between some of the methods. These are shown as CD (\textit{Critical Distance}) plots in Fig.~\ref{Fig.CD}, with one CD plot per performance metric.

The CD plots rank the resampling algorithms from better (lower rank) to worse. The \textit{critical distance} marks the minimum difference between two ranks to be considered statistically significant, drawn as lines connecting those methods whose ranks do not differ more than this.

\begin{figure}[h!]
	\centering
	\includegraphics[width=.7\linewidth]{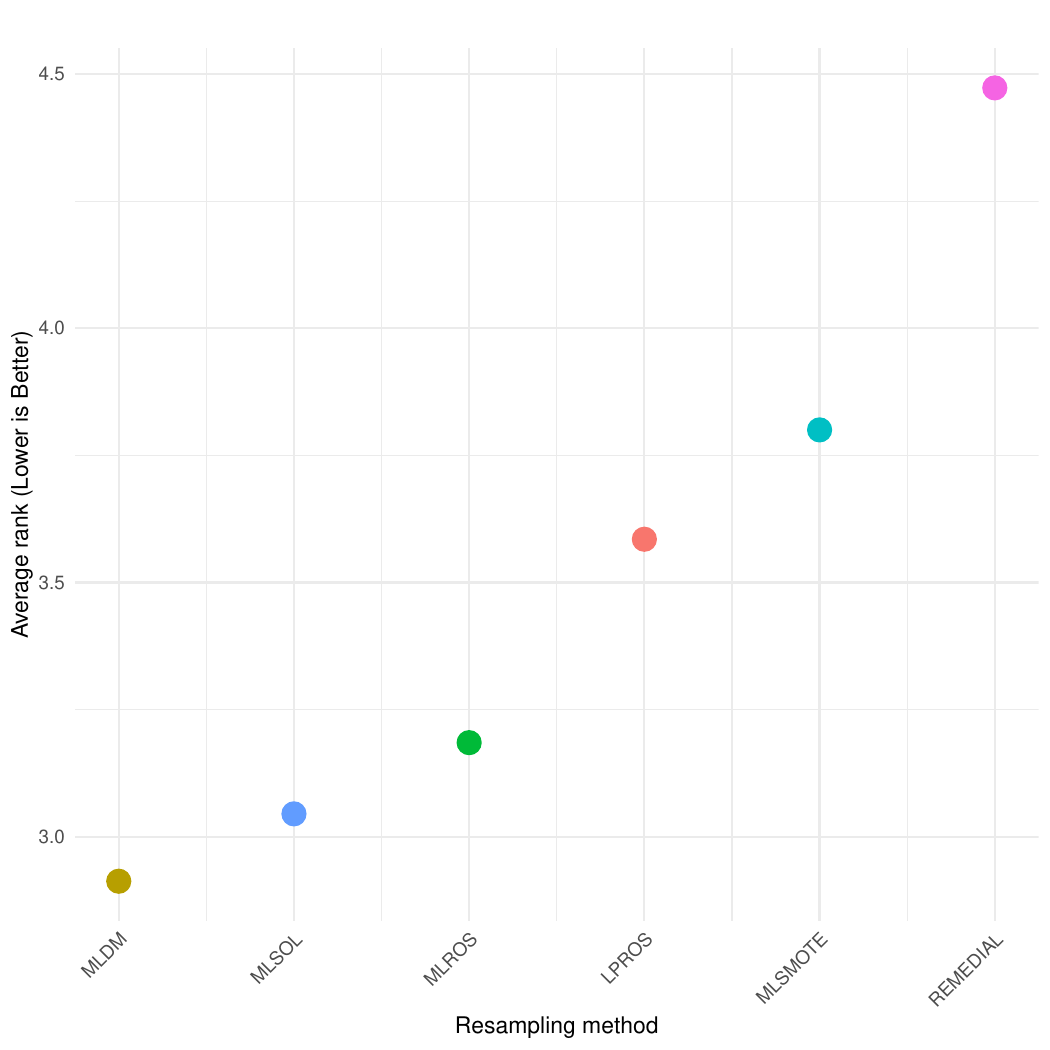}
	\caption{Slope Graph of Methods Ranking. The lower the rank the better performance.}
	\label{Fig.Slope}
\end{figure}

Lastly, to offer a broader view of these results, the average ranking for each resampling algorithm has been obtained and compared with the other in Fig.~\ref{Fig.Slope}.

\subsubsection{Running times}

\begin{figure}[h!]
	\includegraphics[width=\linewidth]{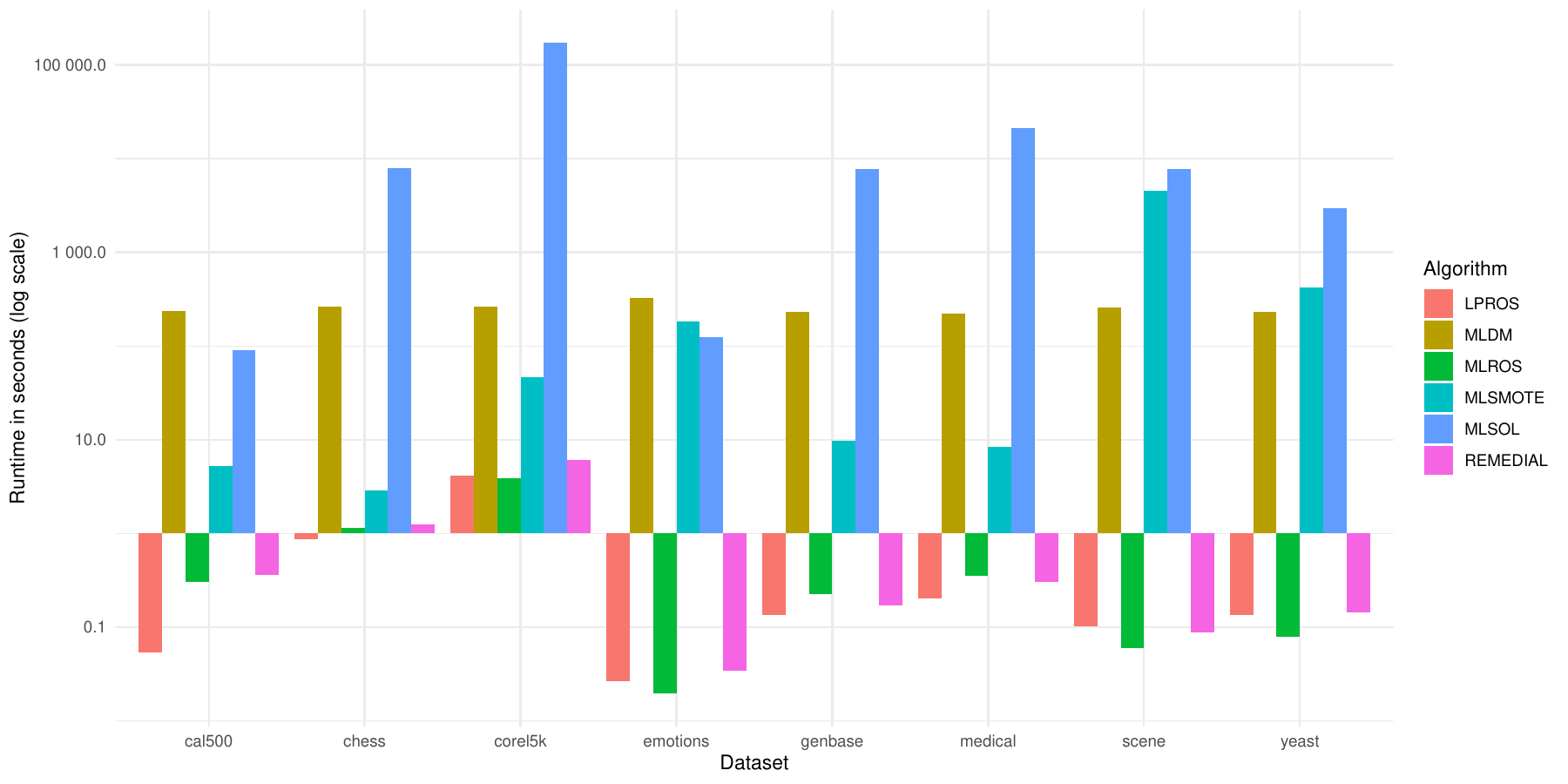}
	\caption{Running times spent by each oversampling algorithm grouped by dataset. }
	\label{Fig.Runtimes}
\end{figure}

In addition to the performance measurements, the time taken by each MOA to process each dataset was also recorded. The collected data is presented in Table~\ref{Tbl.Running}. The plot in Fig.~\ref{Fig.Runtimes} allows a quick comparison of the efficiency of each algorithm. In this case, no statistical tests were performed because the differences are very obvious.

\begin{table*}[h!]
	\centering\footnotesize
	\begin{tabular}{rrrrrrrrr}
		\toprule
		& cal500 & chess & corel5k & emotions & genbase & medical & scene & yeast \\ 
		\midrule
		LPROS & 0.05 & 0.87 & 4.13 & 0.03 & 0.13 & 0.20 & 0.10 & 0.14 \\ 
		MLDM & 237.40 & 262.60 & 264.60 & 322.20 & 231.60 & 220.80 & 257.80 & 232.40 \\ 
		MLROS & 0.30 & 1.15 & 3.85 & 0.02 & 0.22 & 0.35 & 0.06 & 0.08 \\ 
		MLSMOTE & 5.19 & 2.89 & 46.13 & 183.33 & 9.71 & 8.32 & 4\,492.37 & 420.53 \\ 
		MLSOL & 90.50 & 7\,938.83 & 173\,074.20 & 123.63 & 7\,738.66 & 21\,110.38 & 7\,755.57 & 2\,943.30 \\ 
		REMEDIAL & 0.36 & 1.25 & 6.02 & 0.03 & 0.17 & 0.30 & 0.09 & 0.14 \\ 
		\bottomrule
	\end{tabular}
	\caption{Running times spent by each oversampling algorithm in processing each dataset, expressed in seconds.} 
	\label{Tbl.Running}
\end{table*}

\section{Discussion}\label{Sec.Discussion}

In order to draw conclusions, it is necessary to conduct an exhaustive analysis of the reported results. This analysis follows the same structure as previously used, beginning with an examination of imbalance levels, then proceeding to an evaluation of classifier performance, and finally, an assessment of efficiency.

From the data shown in Figure~\ref{Fig.MeanIR}, it is clear that MLDM is able to consistently and significantly reduce the imbalance level of all MLDs. MLSMOTE is close to this behavior, but the imbalance reduction is lower in almost all cases and it seems to have problems with the cal500 dataset. The behavior of MLSOL is also very consistent, although it does not have a significant impact on the global imbalance level of the datasets. MLSOL follows a sophisticated procedure to build the labelset of synthetic samples, taking into account its own metric to determine whether a given sample is considered an outlier, a rare pattern, a borderline case, or a safe case. The labelset is built according to this metric, a seed sample, and a reference sample. A method that does not reduce the global imbalance is not necessarily bad, as long as the new samples help improve the classifiers.

The raw performance values presented in the tables (provided as additional material) are obtained after preprocessing the MLDs with each MOA, training and evaluating the classifiers. These have been grouped and visually presented as radar plots (Figure~\ref{Fig.RadarPlot}) to facilitate this analysis. For each metric (row), the area corresponding to each MOA (column) can be easily compared. The larger the area, the better the performance. It can be seen that MLDM almost always produces the largest area, with the exception of the HL metric. MLSOL and MLROS are very close in some cases. To better illustrate the differences between these algorithms, the CD plots in Figure~\ref{Fig.CD} are generated to represent the output of the Nemenyi statistical test. Some conclusions can be drawn from these plots:

\begin{itemize}
	\item Both MLDM and MLSOL tend to rank first for all metrics, with statistically significant differences from the other MOAs in some cases.

	\item There are no statistically significant differences between MLDM, MLSOL and MLROS. On the contrary, REMEDIAL, LPROS, and sometimes MLSMOTE are worse according to this statistical test.

	\item MLDM is better than MLSOL in three metrics (HL, Micro-F1 and OE), worse in one (Macro-F1) and there is a tie (F1).
\end{itemize}

The average ranking for all metrics, shown in Figure~\ref{Fig.Slope}, confirms that MLDM is the best performer, closely followed by MLSOL, while the other MOAs fall behind.

The third aspect evaluated is the efficiency of the MOAs in preprocessing the MLDs. The running time spent by each MOA on a dataset is shown in Table~\ref{Tbl.Running}. These times are the sum of the time spent in the first phase of the algorithm, i.e. training, and the time spent in generating new samples. Figure~\ref{Fig.Runtimes} is a bar chart of this data with a logarithmic Y-axis. So those cases where less than 1 second is spent appear as downward bars. For most MLDs, the LPROS, MLROS, and REMEDIAL algorithms are very fast. In comparison, methods based on NN search, such as MLSMOTE and MLSOL, take much more time. As can be seen, the time spent by MLDM is nearly constant regardless of the MLD processed. Most of this time is spent on building the diffusion model, while generating the new instances takes less time. For MLDs with more than a hundred instances, MLDM is clearly more efficient than MLSOL, which is almost at par in classification performance.

\section{Conclusions}\label{Sec.Conclusions}

Multilabel learning, whether applied to classification, ranking, or distribution learning, is a complex task. Large imbalances, meaning that some labels are very common while others are rare, make it even more difficult. Resampling algorithms have been proposed as a model-independent approach to alleviate this problem. Oversampling tends to be preferred over undersampling because it does not suffer from potentially useful data loss. Most multilabel oversampling proposals in the literature are based on NN search, with some simpler algorithms that merely clone existing data patterns.

In this paper, we introduced MLDM, a novel oversampling method for multilabel data. It is based on a modern diffusion model instead of NN search. The model is trained on the subset of samples with minority labels, which is always much smaller than the full dataset. The synthetic samples produced by MLDM include not only the features, but also a synthetic labelset, instead of computing it separately as most existing MOAs do. Although MLDM requires a training process, the computational complexity of the new sample generation stage is more efficient than that of other competing proposals because it does not require time-consuming distance computations.

The empirical experiments, conducted on a set of MLDs with different characteristics (amount of samples, number of labels, imbalance levels, etc.), several multilabel classifiers based on diverse approaches (binarization, ensemble, neural networks, NN, etc.), and a variety of performance metrics (sample-based, label-based and ranking-based), have shown that MLDM is very competitive. Although there are no statistically significant differences, MLDM ranks above and is more efficient than excellent MOAs such as MLSOL and MLSMOTE.

An open source implementation of MLDM is already available for anyone to use. We hope that practitioners and researchers will find it useful, and perhaps it will be the seed for other multilabel resampling proposals based on modern generative models, with better performance and efficiency, goals we will try to pursue.

\section*{Acknowledgments}
	
	%The authors would like to thank the anonymous reviewers, whose comments contributed to improving this paper.
		
	The research carried out in this study is part of the project ``Advances in the development of trustworthy AI models to contribute to the adoption and use of responsible AI in healthcare (TAIH)" financed by the Spanish Ministry of Science, Innovation and Universities, code PID2023-149511OB-I00. %Funding for open access charge: Universidad de Jaén

\bibliographystyle{IEEEtran}
\bibliography{IEEEabrv,MLDM}

\begin{thebibliography}{10}
\providecommand{\url}[1]{#1}
\csname url@samestyle\endcsname
\providecommand{\newblock}{\relax}
\providecommand{\bibinfo}[2]{#2}
\providecommand{\BIBentrySTDinterwordspacing}{\spaceskip=0pt\relax}
\providecommand{\BIBentryALTinterwordstretchfactor}{4}
\providecommand{\BIBentryALTinterwordspacing}{\spaceskip=\fontdimen2\font plus
\BIBentryALTinterwordstretchfactor\fontdimen3\font minus
  \fontdimen4\font\relax}
\providecommand{\BIBforeignlanguage}[2]{{%
\expandafter\ifx\csname l@#1\endcsname\relax
\typeout{** WARNING: IEEEtran.bst: No hyphenation pattern has been}%
\typeout{** loaded for the language `#1'. Using the pattern for}%
\typeout{** the default language instead.}%
\else
\language=\csname l@#1\endcsname
\fi
#2}}
\providecommand{\BIBdecl}{\relax}
\BIBdecl

\bibitem{he2009learning}
H.~He and E.~A. Garcia, ``Learning from imbalanced data,'' \emph{IEEE
  Transactions on Knowledge and Data Engineering}, vol.~21, no.~9, pp.
  1263--1284, 2009.

\bibitem{kaur2019A}
H.~Kaur, H.~Pannu, and A.~Malhi, ``A systematic review on imbalanced data
  challenges in machine learning,'' \emph{ACM Computing Surveys (CSUR)},
  vol.~52, pp. 1 -- 36, 2019.

\bibitem{CharteMLC}
F.~Herrera, F.~Charte, A.~J. Rivera, and M.~J. Del~Jesus, \emph{Multilabel
  classification. Problem Analysis, Metrics and Techniques}.\hskip 1em plus
  0.5em minus 0.4em\relax Springer, 2016.

\bibitem{vembu2010label}
S.~Vembu and T.~G{\"a}rtner, ``Label ranking algorithms: A survey,'' in
  \emph{Preference learning}.\hskip 1em plus 0.5em minus 0.4em\relax Springer,
  2010, pp. 45--64.

\bibitem{geng2016label}
X.~Geng, ``Label distribution learning,'' \emph{IEEE Transactions on Knowledge
  and Data Engineering}, vol.~28, no.~7, pp. 1734--1748, 2016.

\bibitem{charte2015addressing}
F.~Charte, A.~J. Rivera, M.~J. {del Jesus}, and F.~Herrera, ``Addressing
  imbalance in multilabel classification: Measures and random resampling
  algorithms,'' \emph{Neurocomputing}, vol. 163, pp. 3--16, 2015.

\bibitem{charte2019dealing}
F.~Charte, A.~J. Rivera, M.~J. del Jesus, and F.~Herrera, ``Dealing with
  difficult minority labels in imbalanced mutilabel data sets,''
  \emph{Neurocomputing}, vol. 326, pp. 39--53, 2019.

\bibitem{maloof2003learning}
M.~A. Maloof, ``Learning when data sets are imbalanced and when costs are
  unequal and unknown,'' in \emph{ICML-2003 workshop on learning from
  imbalanced data sets II}, vol.~2.\hskip 1em plus 0.5em minus 0.4em\relax
  Washington DC, 2003, pp. 2--1.

\bibitem{sun2007cost}
Y.~Sun, M.~S. Kamel, A.~K. Wong, and Y.~Wang, ``Cost-sensitive boosting for
  classification of imbalanced data,'' \emph{Pattern recognition}, vol.~40,
  no.~12, pp. 3358--3378, 2007.

\bibitem{liu2009easyensemble}
T.-Y. Liu, ``Easyensemble and feature selection for imbalance data sets,'' in
  \emph{2009 international joint conference on bioinformatics, systems biology
  and intelligent computing}.\hskip 1em plus 0.5em minus 0.4em\relax IEEE,
  2009, pp. 517--520.

\bibitem{Chawla2002}
N.~V. Chawla, K.~W. Bowyer, L.~O. Hall, and W.~P. Kegelmeyer, ``Smote:
  synthetic minority over-sampling technique,'' \emph{Journal of artificial
  intelligence research}, vol.~16, pp. 321--357, 2002.

\bibitem{Liu2009}
X.-Y. Liu, J.~Wu, and Z.-H. Zhou, ``Exploratory undersampling for
  class-imbalance learning,'' in \emph{2009 IEEE International Conference on
  Systems, Man and Cybernetics}.\hskip 1em plus 0.5em minus 0.4em\relax IEEE,
  2009, pp. 2923--2928.

\bibitem{kingma2013auto}
D.~P. Kingma and M.~Welling, ``Auto-encoding variational bayes,'' \emph{arXiv
  preprint arXiv:1312.6114}, 2013.

\bibitem{goodfellow2014generative}
I.~Goodfellow, J.~Pouget-Abadie, M.~Mirza, B.~Xu, D.~Warde-Farley, S.~Ozair,
  A.~Courville, and Y.~Bengio, ``Generative adversarial nets,'' in
  \emph{Advances in neural information processing systems}, 2014, pp.
  2672--2680.

\bibitem{rezende2015variational}
D.~Rezende and S.~Mohamed, ``Variational inference with normalizing flows,'' in
  \emph{International conference on machine learning}.\hskip 1em plus 0.5em
  minus 0.4em\relax PMLR, 2015, pp. 1530--1538.

\bibitem{sohl2015deep}
J.~Sohl-Dickstein, E.~Weiss, N.~Maheswaranathan, and S.~Ganguli, ``Deep
  unsupervised learning using nonequilibrium thermodynamics,'' in
  \emph{International Conference on Machine Learning}.\hskip 1em plus 0.5em
  minus 0.4em\relax PMLR, 2015, pp. 2256--2265.

\bibitem{vaswani2017attention}
A.~Vaswani, N.~Shazeer, N.~Parmar, J.~Uszkoreit, L.~Jones, A.~N. Gomez,
  {\L}.~Kaiser, and I.~Polosukhin, ``Attention is all you need,'' in
  \emph{Advances in neural information processing systems}, 2017, pp.
  5998--6008.

\bibitem{mlsmote}
F.~Charte, A.~J. Rivera, M.~J. {del Jesus}, and F.~Herrera, ``{MLSMOTE}:
  Approaching imbalanced multilabel learning through synthetic instance
  generation,'' \emph{Knowledge-Based Systems}, vol.~89, pp. 385--397, 2015.

\bibitem{mlsol}
B.~Liu, K.~Blekas, and G.~Tsoumakas, ``Multi-label sampling based on local
  label imbalance,'' \emph{Pattern Recognition}, vol. 122, p. 108294, 2022.

\bibitem{robinson2018multi}
S.~D. Robinson, ``Multi-label classification of contributing causal factors in
  self-reported safety narratives,'' \emph{Safety}, vol.~4, p.~30, 2018.

\bibitem{dai2018novel}
O.~E. Dai, B.~Demir, B.~Sankur, and L.~Bruzzone, ``A novel system for
  content-based retrieval of single and multi-label high-dimensional remote
  sensing images,'' \emph{IEEE Journal of Selected Topics in Applied Earth
  Observations and Remote Sensing}, no.~99, pp. 1--18, 2018.

\bibitem{liu2018integrated}
T.~Liu, L.~Chen, and X.~Pan, ``An integrated multi-label classifier with
  chemical-chemical interactions for prediction of chemical toxicity effects,''
  \emph{Combinatorial chemistry \& high throughput screening}, vol.~21, no.~6,
  pp. 403--410, 2018.

\bibitem{stackexchess}
F.~Charte, A.~J. Rivera, M.~J. del Jesus, and F.~Herrera, ``Quinta: A question
  tagging assistant to improve the answering ratio in electronic forums,'' in
  \emph{EUROCON 2015 - International Conference on Computer as a Tool
  (EUROCON), IEEE}, Sept 2015, pp. 1--6.

\bibitem{Zhang:2013}
M.~Zhang and Z.~Zhou, ``A review on multi-label learning algorithms,''
  \emph{IEEE Transactions on Knowledge and Data Engineering}, vol.~26, no.~8,
  pp. 1819--1837, Aug 2014.

\bibitem{han2023survey}
M.~Han, H.~Wu, Z.~Chen, M.~Li, and X.~Zhang, ``A survey of multi-label
  classification based on supervised and semi-supervised learning,''
  \emph{International Journal of Machine Learning and Cybernetics}, vol.~14,
  no.~3, pp. 697--724, 2023.

\bibitem{Alberto:2013}
A.~Fern\'andez, V.~L\'opez, M.~Galar, M.~J. del Jesus, and F.~Herrera,
  ``Analysing the classification of imbalanced data-sets with multiple classes:
  Binarization techniques and ad-hoc approaches,'' \emph{Knowl. Based Systems},
  vol.~42, pp. 97 -- 110, 2013.

\bibitem{Godbole}
S.~Godbole and S.~Sarawagi, ``{Discriminative Methods for Multi-Labeled
  Classification},'' in \emph{Advances in Knowledge Discovery and Data Mining},
  vol. 3056, 2004, pp. 22--30.

\bibitem{Charte:NeucomDifficultLabels}
F.~Charte, A.~J. Rivera, M.~J. del Jesus, and F.~Herrera, ``Dealing with
  difficult minority labels in imbalanced mutilabel data sets,''
  \emph{Neurocomputing}, vol. 326, pp. 39--53, 2019.

\bibitem{kotsiantis2006handling}
S.~Kotsiantis, D.~Kanellopoulos, P.~Pintelas \emph{et~al.}, ``Handling
  imbalanced datasets: A review,'' \emph{GESTS international transactions on
  computer science and engineering}, vol.~30, no.~1, pp. 25--36, 2006.

\bibitem{mohammed2020machine}
R.~Mohammed, J.~Rawashdeh, and M.~Abdullah, ``Machine learning with
  oversampling and undersampling techniques: overview study and experimental
  results,'' in \emph{2020 11th international conference on information and
  communication systems (ICICS)}.\hskip 1em plus 0.5em minus 0.4em\relax IEEE,
  2020, pp. 243--248.

\bibitem{Batista2004}
G.~E. Batista, R.~C. Prati, and M.~C. Monard, ``A study of the behavior of
  several methods for balancing machine learning training data,'' in \emph{ACM
  SIGKDD explorations newsletter}, vol.~6, no.~1.\hskip 1em plus 0.5em minus
  0.4em\relax ACM, 2004, pp. 20--29.

\bibitem{remedial}
F.~Charte, A.~Rivera~Rivas, M.~J. Del~Jesus, and F.~Herrera, ``Resampling
  multilabel datasets by decoupling highly imbalanced labels,'' in
  \emph{Lecture Notes in Artificial Intelligence (Subseries of Lecture Notes in
  Computer Science)}, vol. 9121, 06 2015.

\bibitem{rhwrsmt}
F.~Charte, A.~J. Rivera, M.~J. {del Jesus}, and F.~Herrera, ``{REMEDIAL-HwR}:
  Tackling multilabel imbalance through label decoupling and data resampling
  hybridization,'' \emph{Neurocomputing}, vol. 326-327, pp. 110--122, 2019.

\bibitem{potdar2017comparative}
K.~Potdar, T.~S. Pardawala, and C.~D. Pai, ``A comparative study of categorical
  variable encoding techniques for neural network classifiers,''
  \emph{International journal of computer applications}, vol. 175, no.~4, pp.
  7--9, 2017.

\bibitem{peterson2020ordered}
R.~A. Peterson and J.~E. Cavanaugh, ``Ordered quantile normalization: a
  semiparametric transformation built for the cross-validation era,''
  \emph{Journal of applied statistics}, 2020.

\bibitem{hoogeboom2021argmax}
E.~Hoogeboom, D.~Nielsen, P.~Jaini, P.~Forr{\'e}, and M.~Welling, ``Argmax
  flows and multinomial diffusion: Learning categorical distributions,''
  \emph{Advances in Neural Information Processing Systems}, vol.~34, pp.
  12\,454--12\,465, 2021.

\bibitem{kotelnikov2023tabddpm}
A.~Kotelnikov, D.~Baranchuk, I.~Rubachev, and A.~Babenko, ``Tabddpm: Modelling
  tabular data with diffusion models,'' in \emph{International Conference on
  Machine Learning}.\hskip 1em plus 0.5em minus 0.4em\relax PMLR, 2023, pp.
  17\,564--17\,579.

\bibitem{LI-MLC}
F.~Charte, A.~J. Rivera, M.~J. del Jesus, and F.~Herrera, ``{LI-MLC: A label
  inference methodology for addressing high dimensionality in the label space
  for multilabel classification},'' \emph{IEEE transactions on neural networks
  and learning systems}, vol.~25, no.~10, pp. 1842--1854, 2014.

\bibitem{cometaml}
F.~Charte, A.~J. Rivera, D.~Charte, M.~J. del Jesus, and F.~Herrera, ``{Tips,
  guidelines and tools for managing multi-label datasets: The mldr.datasets R
  package and the Cometa data repository},'' \emph{Neurocomputing}, 2018.

\bibitem{cal500}
D.~Turnbull, L.~Barrington, D.~Torres, and G.~Lanckriet, ``Semantic annotation
  and retrieval of music and sound effects,'' \emph{Audio, Speech, and Language
  Processing, IEEE Transactions on}, vol.~16, no.~2, pp. 467--476, 2008.

\bibitem{corel5k}
P.~Duygulu, K.~Barnard, J.~F. de~Freitas, and D.~A. Forsyth, ``Object
  recognition as machine translation: Learning a lexicon for a fixed image
  vocabulary,'' in \emph{Computer Vision, ECCV 2002}, ser. LNCS, 2002, vol.
  2353, pp. 97--112.

\bibitem{emotions}
A.~Wieczorkowska, P.~Synak, and Z.~Ra'{s}, ``Multi-label classification of
  emotions in music,'' in \emph{Intelligent Information Processing and Web
  Mining}, 2006, vol.~35, ch.~30, pp. 307--315.

\bibitem{genbase}
S.~Diplaris, G.~Tsoumakas, P.~Mitkas, and I.~Vlahavas, ``Protein classification
  with multiple algorithms,'' in \emph{Proc. 10th Panhellenic Conference on
  Informatics, Volos, Greece, PCI05}, 2005, pp. 448--456.

\bibitem{medical}
K.~Crammer, M.~Dredze, K.~Ganchev, P.~P. Talukdar, and S.~Carroll, ``Automatic
  code assignment to medical text,'' in \emph{Proc. Workshop on Biological,
  Translational, and Clinical Language Processing, Prague, Czech Republic,
  BioNLP07}, 2007, pp. 129--136.

\bibitem{scene}
M.~Boutell, J.~Luo, X.~Shen, and C.~Brown, ``Learning multi-label scene
  classification,'' \emph{Pattern Recognition}, vol.~37, no.~9, pp. 1757--1771,
  2004.

\bibitem{yeast}
A.~Elisseeff and J.~Weston, ``A kernel method for multi-labelled
  classification,'' in \emph{Advances in Neural Information Processing
  Systems}, vol.~14, 2001, pp. 681--687.

\bibitem{japkowicz2002class}
N.~Japkowicz and S.~Stephen, ``The class imbalance problem: A systematic
  study,'' \emph{Intelligent data analysis}, vol.~6, no.~5, pp. 429--449, 2002.

\bibitem{lopez2013insight}
V.~L{\'o}pez, A.~Fern{\'a}ndez, S.~Garc{\'\i}a, V.~Palade, and F.~Herrera, ``An
  insight into classification with imbalanced data: Empirical results and
  current trends on using data intrinsic characteristics,'' \emph{Information
  sciences}, vol. 250, pp. 113--141, 2013.

\bibitem{br.lp}
G.~Tsoumakas, I.~Katakis, and I.~Vlahavas, \emph{Mining Multi-label
  Data}.\hskip 1em plus 0.5em minus 0.4em\relax Boston, MA: Springer US, 2010,
  pp. 667--685.

\bibitem{bpmll}
Z.~Z.-H. Zhang, Min-Ling., ``Multi-label neural networks with applications to
  functional genomics and text categorization,'' \emph{IEEE Transactions on
  Knowledge and Data Engineering}, vol.~18, pp. 1338--1351, 2006.

\bibitem{mlknn}
M.-L. Zhang and Z.-H. Zhou, ``Ml-knn: A lazy learning approach to multi-label
  learning,'' \emph{Pattern Recogn.}, vol.~40, no.~7, pp. 2038--2048, 2007.

\bibitem{homer}
G.~Tsoumakas, I.~Katakis, and I.~Vlahavas, ``Effective and efficient multilabel
  classification in domains with large number of labels,'' in \emph{Proc.
  ECML/PKDD 2008 Workshop on Mining Multidimensional Data (MMD'08)}, 2008.

\bibitem{mldr.resampling}
A.~J. Rivera, M.~A. D{\'a}vila, M.~J. del Jesus, and F.~Charte,
  ``mldr.resampling: Efficient reference implementations of multilabel
  resampling algorithms,'' \emph{Neurocomputing}, vol. 559, p. 126806, 2023.

\bibitem{MULAN}
G.~Tsoumakas, E.~Spyromitros-Xioufis, J.~Vilcek, and I.~Vlahavas, ``Mulan: A
  java library for multi-label learning,'' \emph{The Journal of Machine
  Learning Research}, vol.~12, pp. 2411--2414, 2011.

\bibitem{demvsar2006statistical}
J.~Dem{\v{s}}ar, ``Statistical comparisons of classifiers over multiple data
  sets,'' \emph{The Journal of Machine learning research}, vol.~7, pp. 1--30,
  2006.

\end{thebibliography}
	
\clearpage
\section*{Additional material}\label{Sec.Results}
The following tables provide full results for each metric, resampling method, classifier and dataset.
\begin{table*}[h!]
	\centering\renewcommand{\arraystretch}{0.9}
	\tiny
	\begin{tabular}{llcccccccc}
		\toprule
		\textbf{Classifier} & \textbf{Resampling} & cal500 & chess & corel5k & emotions & genbase & medical & scene & yeast \\ 
		\midrule
		BPMLL & None & 0.452$\pm$0.010 & 0.016$\pm$0.002 & 0.024$\pm$0.001 & 0.654$\pm$0.013 & 0.039$\pm$0.044 & \textbf{0.053$\pm$0.031} & 0.472$\pm$0.042 & 0.623$\pm$0.023 \\ 
		& LPROS & 0.456$\pm$0.019 & 0.014$\pm$0.002 & 0.021$\pm$0.001 & 0.647$\pm$0.040 & 0.052$\pm$0.048 & 0.033$\pm$0.030 & 0.450$\pm$0.024 & 0.627$\pm$0.022 \\ 
		& MLDM & \textbf{0.464$\pm$0.015} & \textbf{0.017$\pm$0.002} & 0.024$\pm$0.001 & 0.654$\pm$0.028 & 0.065$\pm$0.023 & 0.024$\pm$0.033 & \textbf{0.518$\pm$0.057} & \textbf{0.632$\pm$0.029} \\ 
		& MLROS & 0.452$\pm$0.020 & 0.015$\pm$0.004 & 0.025$\pm$0.002 & \textbf{0.663$\pm$0.021} & \textbf{0.089$\pm$0.020} & 0.052$\pm$0.030 & 0.508$\pm$0.041 & 0.621$\pm$0.019 \\ 
		& MLSMOTE & 0.463$\pm$0.017 & 0.014$\pm$0.005 & 0.015$\pm$0.002 & 0.620$\pm$0.039 & 0.072$\pm$0.028 & 0.035$\pm$0.032 & 0.470$\pm$0.051 & 0.614$\pm$0.014 \\ 
		& MLSOL & 0.456$\pm$0.014 & \textbf{0.017$\pm$0.001} & \textbf{0.026$\pm$0.002} & 0.658$\pm$0.044 & 0.073$\pm$0.047 & 0.046$\pm$0.042 & 0.459$\pm$0.054 & 0.629$\pm$0.018 \\ 
		& REMEDIAL & 0.408$\pm$0.011 & 0.007$\pm$0.007 & 0.007$\pm$0.001 & 0.557$\pm$0.026 & 0.065$\pm$0.041 & 0.064$\pm$0.004 & 0.499$\pm$0.020 & 0.589$\pm$0.009 \\ 
		\cmidrule{1-2}
		BR-J48 & None & 0.347$\pm$0.016 & 0.249$\pm$0.017 & 0.088$\pm$0.008 & \textbf{0.552$\pm$0.021} & 0.984$\pm$0.020 & 0.771$\pm$0.021 & \textbf{0.580$\pm$0.019} & 0.554$\pm$0.020 \\ 
		& LPROS & 0.347$\pm$0.016 & 0.241$\pm$0.031 & \textbf{0.102$\pm$0.007} & 0.540$\pm$0.043 & 0.936$\pm$0.110 & \textbf{0.781$\pm$0.021} & 0.566$\pm$0.018 & 0.528$\pm$0.011 \\ 
		& MLDM & 0.336$\pm$0.023 & \textbf{0.254$\pm$0.019} & 0.085$\pm$0.009 & 0.548$\pm$0.026 & \textbf{0.989$\pm$0.010} & 0.772$\pm$0.013 & 0.545$\pm$0.023 & \textbf{0.565$\pm$0.016} \\ 
		& MLROS & 0.339$\pm$0.011 & 0.252$\pm$0.016 & 0.097$\pm$0.006 & 0.510$\pm$0.026 & \textbf{0.989$\pm$0.022} & 0.766$\pm$0.023 & 0.579$\pm$0.035 & 0.524$\pm$0.025 \\ 
		& MLSMOTE & 0.314$\pm$0.007 & 0.236$\pm$0.022 & 0.085$\pm$0.006 & 0.520$\pm$0.023 & 0.929$\pm$0.096 & 0.766$\pm$0.014 & 0.565$\pm$0.022 & 0.525$\pm$0.023 \\ 
		& MLSOL & \textbf{0.359$\pm$0.015} & 0.243$\pm$0.024 & 0.096$\pm$0.007 & 0.525$\pm$0.023 & 0.934$\pm$0.094 & 0.775$\pm$0.016 & 0.571$\pm$0.017 & 0.537$\pm$0.020 \\ 
		& REMEDIAL & 0.154$\pm$0.022 & 0.080$\pm$0.029 & 0.034$\pm$0.004 & 0.317$\pm$0.026 & 0.932$\pm$0.105 & 0.700$\pm$0.021 & 0.555$\pm$0.035 & 0.484$\pm$0.068 \\ 
		\cmidrule{1-2}
		HOMER & None & \textbf{0.395$\pm$0.011} & 0.247$\pm$0.015 & 0.149$\pm$0.010 & 0.546$\pm$0.026 & 0.985$\pm$0.017 & \textbf{0.784$\pm$0.018} & 0.551$\pm$0.024 & 0.552$\pm$0.032 \\ 
		& LPROS & 0.393$\pm$0.017 & 0.241$\pm$0.007 & 0.143$\pm$0.007 & 0.528$\pm$0.025 & 0.943$\pm$0.096 & 0.746$\pm$0.020 & 0.555$\pm$0.011 & 0.526$\pm$0.015 \\ 
		& MLDM & 0.391$\pm$0.014 & 0.258$\pm$0.010 & 0.156$\pm$0.007 & 0.545$\pm$0.034 & 0.987$\pm$0.011 & 0.781$\pm$0.016 & 0.548$\pm$0.020 & \textbf{0.562$\pm$0.027} \\ 
		& MLROS & 0.349$\pm$0.019 & 0.238$\pm$0.017 & \textbf{0.157$\pm$0.008} & 0.523$\pm$0.027 & \textbf{0.989$\pm$0.013} & 0.751$\pm$0.018 & 0.537$\pm$0.011 & 0.530$\pm$0.019 \\ 
		& MLSMOTE & 0.335$\pm$0.013 & 0.253$\pm$0.017 & 0.148$\pm$0.006 & 0.513$\pm$0.026 & 0.920$\pm$0.117 & 0.774$\pm$0.016 & 0.548$\pm$0.024 & 0.535$\pm$0.014 \\ 
		& MLSOL & 0.383$\pm$0.010 & \textbf{0.272$\pm$0.007} & 0.145$\pm$0.005 & \textbf{0.547$\pm$0.023} & 0.932$\pm$0.101 & 0.758$\pm$0.012 & \textbf{0.561$\pm$0.015} & 0.531$\pm$0.023 \\ 
		& REMEDIAL & 0.392$\pm$0.011 & 0.203$\pm$0.017 & 0.115$\pm$0.005 & 0.438$\pm$0.065 & 0.937$\pm$0.103 & 0.734$\pm$0.045 & 0.550$\pm$0.025 & 0.557$\pm$0.043 \\ 
		\cmidrule{1-2}
		LP-J48 & None & 0.328$\pm$0.014 & 0.139$\pm$0.008 & 0.102$\pm$0.010 & 0.534$\pm$0.041 & 0.978$\pm$0.023 & 0.765$\pm$0.039 & 0.588$\pm$0.034 & 0.506$\pm$0.018 \\ 
		& LPROS & 0.328$\pm$0.014 & 0.147$\pm$0.018 & \textbf{0.108$\pm$0.006} & \textbf{0.559$\pm$0.008} & 0.936$\pm$0.099 & 0.754$\pm$0.023 & 0.578$\pm$0.024 & 0.500$\pm$0.020 \\ 
		& MLDM & 0.322$\pm$0.012 & \textbf{0.154$\pm$0.011} & \textbf{0.108$\pm$0.008} & 0.527$\pm$0.025 & \textbf{0.985$\pm$0.009} & \textbf{0.766$\pm$0.041} & 0.583$\pm$0.030 & 0.504$\pm$0.007 \\ 
		& MLROS & 0.331$\pm$0.017 & 0.150$\pm$0.016 & 0.102$\pm$0.001 & 0.529$\pm$0.015 & \textbf{0.985$\pm$0.030} & 0.750$\pm$0.025 & 0.592$\pm$0.018 & \textbf{0.513$\pm$0.017} \\ 
		& MLSMOTE & 0.304$\pm$0.010 & 0.135$\pm$0.013 & 0.104$\pm$0.006 & 0.522$\pm$0.029 & 0.930$\pm$0.109 & 0.753$\pm$0.039 & \textbf{0.601$\pm$0.023} & 0.498$\pm$0.014 \\ 
		& MLSOL & \textbf{0.334$\pm$0.011} & \textbf{0.154$\pm$0.013} & 0.107$\pm$0.005 & 0.531$\pm$0.027 & 0.920$\pm$0.114 & 0.765$\pm$0.019 & 0.598$\pm$0.020 & 0.497$\pm$0.018 \\ 
		& REMEDIAL & 0.231$\pm$0.017 & 0.033$\pm$0.012 & 0.025$\pm$0.005 & 0.505$\pm$0.027 & 0.928$\pm$0.108 & 0.670$\pm$0.053 & 0.572$\pm$0.021 & 0.355$\pm$0.034 \\ 
		\cmidrule{1-2}
		MLkNN & None & 0.320$\pm$0.010 & 0.046$\pm$0.018 & 0.016$\pm$0.003 & 0.629$\pm$0.014 & 0.953$\pm$0.041 & 0.585$\pm$0.016 & 0.679$\pm$0.016 & 0.614$\pm$0.029 \\ 
		& LPROS & 0.320$\pm$0.010 & \textbf{0.064$\pm$0.018} & 0.024$\pm$0.006 & 0.582$\pm$0.033 & 0.879$\pm$0.161 & 0.574$\pm$0.052 & 0.668$\pm$0.013 & 0.598$\pm$0.029 \\ 
		& MLDM & 0.326$\pm$0.012 & 0.052$\pm$0.022 & 0.017$\pm$0.002 & \textbf{0.639$\pm$0.013} & 0.955$\pm$0.009 & 0.587$\pm$0.017 & 0.663$\pm$0.035 & \textbf{0.616$\pm$0.031} \\ 
		& MLROS & 0.326$\pm$0.010 & 0.044$\pm$0.012 & 0.028$\pm$0.005 & 0.607$\pm$0.018 & \textbf{0.961$\pm$0.028} & 0.498$\pm$0.028 & 0.673$\pm$0.011 & 0.531$\pm$0.033 \\ 
		& MLSMOTE & 0.276$\pm$0.006 & 0.030$\pm$0.011 & 0.018$\pm$0.002 & 0.607$\pm$0.027 & 0.864$\pm$0.183 & 0.583$\pm$0.016 & 0.682$\pm$0.022 & 0.590$\pm$0.021 \\ 
		& MLSOL & \textbf{0.338$\pm$0.012} & 0.059$\pm$0.009 & \textbf{0.035$\pm$0.005} & 0.630$\pm$0.009 & 0.854$\pm$0.202 & \textbf{0.608$\pm$0.037} & \textbf{0.709$\pm$0.023} & 0.611$\pm$0.021 \\ 
		& REMEDIAL & 0.088$\pm$0.010 & 0.005$\pm$0.004 & 0.012$\pm$0.003 & 0.371$\pm$0.021 & 0.877$\pm$0.168 & 0.466$\pm$0.060 & 0.655$\pm$0.031 & 0.477$\pm$0.064 \\ 
		\bottomrule
	\end{tabular}
	\caption{Average classification performance as measured by the sample-based F1 metric (the higher the better). The best values for each classifier/dataset quadrant are highlighted in bold.} 
\end{table*}

% latex table generated in R 4.2.2 by xtable 1.8-4 package
% Tue Jul  2 12:12:53 2024
\begin{table*}[h!]
	\centering\renewcommand{\arraystretch}{0.9}
	\tiny
	\begin{tabular}{llcccccccc}
		\toprule
		\textbf{Classifier} & \textbf{Resampling} & cal500 & chess & corel5k & emotions & genbase & medical & scene & yeast \\ 
		\midrule
		BPMLL & None & 0.213$\pm$0.009 & 0.021$\pm$0.004 & 0.139$\pm$0.020 & 0.674$\pm$0.010 & 0.141$\pm$0.059 & 0.226$\pm$0.129 & 0.545$\pm$0.022 & 0.438$\pm$0.015 \\ 
		& LPROS & 0.215$\pm$0.021 & 0.024$\pm$0.008 & 0.124$\pm$0.012 & 0.653$\pm$0.043 & \textbf{0.192$\pm$0.082} & 0.158$\pm$0.155 & 0.507$\pm$0.018 & \textbf{0.453$\pm$0.011} \\ 
		& MLDM & \textbf{0.227$\pm$0.014} & 0.022$\pm$0.006 & 0.141$\pm$0.010 & 0.665$\pm$0.032 & 0.111$\pm$0.025 & 0.242$\pm$0.134 & \textbf{0.569$\pm$0.035} & 0.443$\pm$0.012 \\ 
		& MLROS & 0.220$\pm$0.019 & \textbf{0.025$\pm$0.014} & 0.147$\pm$0.010 & \textbf{0.678$\pm$0.019} & 0.157$\pm$0.052 & 0.199$\pm$0.118 & 0.561$\pm$0.042 & 0.436$\pm$0.014 \\ 
		& MLSMOTE & 0.196$\pm$0.014 & 0.020$\pm$0.001 & 0.071$\pm$0.013 & 0.656$\pm$0.039 & 0.124$\pm$0.056 & 0.180$\pm$0.137 & 0.541$\pm$0.040 & 0.436$\pm$0.005 \\ 
		& MLSOL & 0.226$\pm$0.008 & 0.024$\pm$0.009 & \textbf{0.156$\pm$0.016} & 0.673$\pm$0.042 & 0.162$\pm$0.068 & \textbf{0.296$\pm$0.067} & 0.532$\pm$0.018 & 0.445$\pm$0.017 \\ 
		& REMEDIAL & 0.186$\pm$0.005 & 0.017$\pm$0.002 & 0.049$\pm$0.010 & 0.617$\pm$0.021 & 0.168$\pm$0.120 & 0.156$\pm$0.034 & 0.547$\pm$0.014 & 0.411$\pm$0.025 \\ 
		\cmidrule{1-2}
		BR-J48 & None & 0.148$\pm$0.015 & \textbf{0.325$\pm$0.023} & \textbf{0.196$\pm$0.014} & \textbf{0.588$\pm$0.006} & 0.870$\pm$0.159 & \textbf{0.661$\pm$0.061} & \textbf{0.636$\pm$0.013} & 0.381$\pm$0.008 \\ 
		& LPROS & 0.148$\pm$0.015 & 0.308$\pm$0.026 & 0.192$\pm$0.011 & 0.583$\pm$0.039 & 0.839$\pm$0.141 & 0.628$\pm$0.045 & 0.630$\pm$0.009 & 0.397$\pm$0.006 \\ 
		& MLDM & 0.133$\pm$0.012 & 0.288$\pm$0.029 & 0.185$\pm$0.020 & 0.585$\pm$0.015 & \textbf{0.933$\pm$0.073} & 0.634$\pm$0.067 & 0.612$\pm$0.015 & 0.387$\pm$0.008 \\ 
		& MLROS & 0.159$\pm$0.010 & 0.315$\pm$0.027 & 0.190$\pm$0.016 & 0.560$\pm$0.027 & 0.922$\pm$0.173 & 0.654$\pm$0.069 & 0.634$\pm$0.028 & 0.390$\pm$0.011 \\ 
		& MLSMOTE & 0.136$\pm$0.010 & 0.318$\pm$0.024 & 0.178$\pm$0.015 & 0.568$\pm$0.014 & 0.787$\pm$0.169 & 0.627$\pm$0.050 & 0.625$\pm$0.017 & 0.390$\pm$0.011 \\ 
		& MLSOL & \textbf{0.169$\pm$0.017} & 0.318$\pm$0.021 & 0.195$\pm$0.014 & 0.570$\pm$0.025 & 0.818$\pm$0.186 & 0.632$\pm$0.065 & 0.626$\pm$0.008 & \textbf{0.401$\pm$0.011} \\ 
		& REMEDIAL & 0.075$\pm$0.017 & 0.289$\pm$0.016 & 0.188$\pm$0.014 & 0.384$\pm$0.042 & 0.829$\pm$0.182 & 0.626$\pm$0.055 & 0.621$\pm$0.028 & 0.267$\pm$0.015 \\ 
		\cmidrule{1-2}
		HOMER & None & 0.173$\pm$0.007 & 0.275$\pm$0.018 & \textbf{0.193$\pm$0.016} & 0.585$\pm$0.018 & 0.875$\pm$0.141 & \textbf{0.636$\pm$0.048} & 0.612$\pm$0.013 & 0.398$\pm$0.013 \\ 
		& LPROS & 0.169$\pm$0.011 & 0.263$\pm$0.016 & 0.171$\pm$0.012 & 0.579$\pm$0.023 & 0.838$\pm$0.137 & 0.592$\pm$0.053 & \textbf{0.619$\pm$0.009} & 0.393$\pm$0.016 \\ 
		& MLDM & 0.161$\pm$0.008 & 0.269$\pm$0.025 & 0.184$\pm$0.013 & 0.575$\pm$0.028 & \textbf{0.907$\pm$0.061} & 0.618$\pm$0.039 & 0.603$\pm$0.016 & \textbf{0.402$\pm$0.008} \\ 
		& MLROS & 0.172$\pm$0.010 & 0.270$\pm$0.021 & 0.188$\pm$0.015 & 0.567$\pm$0.025 & 0.927$\pm$0.102 & 0.616$\pm$0.070 & 0.603$\pm$0.012 & 0.392$\pm$0.015 \\ 
		& MLSMOTE & 0.152$\pm$0.004 & 0.284$\pm$0.024 & 0.174$\pm$0.012 & 0.558$\pm$0.009 & 0.771$\pm$0.182 & 0.609$\pm$0.043 & 0.606$\pm$0.020 & 0.399$\pm$0.006 \\ 
		& MLSOL & \textbf{0.180$\pm$0.013} & \textbf{0.288$\pm$0.015} & 0.189$\pm$0.021 & \textbf{0.588$\pm$0.031} & 0.825$\pm$0.172 & 0.612$\pm$0.059 & 0.600$\pm$0.004 & 0.395$\pm$0.014 \\ 
		& REMEDIAL & 0.178$\pm$0.015 & \textbf{0.288$\pm$0.026} & 0.192$\pm$0.014 & 0.465$\pm$0.040 & 0.844$\pm$0.168 & 0.612$\pm$0.049 & 0.609$\pm$0.027 & 0.365$\pm$0.037 \\ 
		\cmidrule{1-2}
		LP-J48 & None & 0.156$\pm$0.011 & 0.174$\pm$0.018 & 0.121$\pm$0.014 & 0.556$\pm$0.033 & 0.851$\pm$0.102 & 0.598$\pm$0.047 & 0.592$\pm$0.033 & 0.374$\pm$0.013 \\ 
		& LPROS & 0.156$\pm$0.011 & 0.178$\pm$0.008 & 0.130$\pm$0.018 & \textbf{0.575$\pm$0.007} & 0.828$\pm$0.147 & \textbf{0.590$\pm$0.048} & 0.586$\pm$0.020 & 0.373$\pm$0.019 \\ 
		& MLDM & 0.159$\pm$0.005 & 0.172$\pm$0.030 & 0.116$\pm$0.022 & 0.551$\pm$0.022 & \textbf{0.890$\pm$0.055} & 0.581$\pm$0.086 & 0.587$\pm$0.027 & 0.370$\pm$0.013 \\ 
		& MLROS & 0.156$\pm$0.015 & 0.185$\pm$0.018 & 0.117$\pm$0.013 & 0.560$\pm$0.018 & \textbf{0.890$\pm$0.121} & 0.584$\pm$0.039 & 0.598$\pm$0.017 & \textbf{0.386$\pm$0.014} \\ 
		& MLSMOTE & 0.144$\pm$0.012 & 0.207$\pm$0.013 & 0.127$\pm$0.019 & 0.548$\pm$0.026 & 0.776$\pm$0.129 & 0.555$\pm$0.065 & 0.605$\pm$0.018 & 0.375$\pm$0.014 \\ 
		& MLSOL & \textbf{0.164$\pm$0.013} & 0.196$\pm$0.011 & 0.125$\pm$0.005 & 0.547$\pm$0.026 & 0.783$\pm$0.210 & 0.577$\pm$0.038 & \textbf{0.606$\pm$0.019} & 0.371$\pm$0.020 \\ 
		& REMEDIAL & 0.136$\pm$0.014 & \textbf{0.242$\pm$0.020 }& \textbf{0.168$\pm$0.018} & 0.518$\pm$0.021 & 0.804$\pm$0.219 & 0.560$\pm$0.044 & 0.585$\pm$0.021 & 0.277$\pm$0.017 \\ 
		\cmidrule{1-2}
		MLkNN & None & 0.089$\pm$0.012 & 0.277$\pm$0.020 & 0.185$\pm$0.016 & 0.642$\pm$0.025 & 0.748$\pm$0.151 & 0.531$\pm$0.075 & 0.734$\pm$0.011 & 0.370$\pm$0.004 \\ 
		& LPROS & 0.089$\pm$0.012 & 0.266$\pm$0.019 & 0.176$\pm$0.015 & 0.610$\pm$0.029 & 0.741$\pm$0.142 & 0.506$\pm$0.070 & 0.720$\pm$0.010 & 0.395$\pm$0.014 \\ 
		& MLDM & 0.091$\pm$0.015 & 0.274$\pm$0.025 & 0.182$\pm$0.015 & \textbf{0.653$\pm$0.020} & 0.741$\pm$0.049 & 0.527$\pm$0.072 & 0.724$\pm$0.025 & 0.381$\pm$0.011 \\ 
		& MLROS & 0.102$\pm$0.011 & 0.276$\pm$0.019 & 0.185$\pm$0.016 & 0.618$\pm$0.005 & \textbf{0.754$\pm$0.105} & 0.477$\pm$0.087 & 0.723$\pm$0.008 & 0.349$\pm$0.013 \\ 
		& MLSMOTE & 0.098$\pm$0.009 & 0.275$\pm$0.019 & 0.182$\pm$0.015 & 0.642$\pm$0.020 & 0.697$\pm$0.080 & 0.530$\pm$0.072 & 0.732$\pm$0.016 & 0.394$\pm$0.012 \\ 
		& MLSOL & \textbf{0.111$\pm$0.016} & \textbf{0.279$\pm$0.025} & \textbf{0.189$\pm$0.015} & 0.643$\pm$0.010 & 0.707$\pm$0.172 & \textbf{0.543$\pm$0.081} & \textbf{0.740$\pm$0.014} & \textbf{0.409$\pm$0.012} \\ 
		& REMEDIAL & 0.048$\pm$0.013 & 0.272$\pm$0.020 & 0.182$\pm$0.016 & 0.370$\pm$0.037 & 0.716$\pm$0.131 & 0.505$\pm$0.087 & 0.718$\pm$0.022 & 0.253$\pm$0.010 \\ 
		\bottomrule
	\end{tabular}
	\caption{Average classification performance as measured by the label-based macro-averaged F1 metric (the higher the better). The best values for each classifier/dataset quadrant are highlighted in bold.} 
\end{table*}

% latex table generated in R 4.2.2 by xtable 1.8-4 package
% Tue Jul  2 12:12:53 2024
\begin{table*}[h!]
	\centering\renewcommand{\arraystretch}{0.9}
	\tiny
	\begin{tabular}{llcccccccc}
		\toprule
		\textbf{Classifier} & \textbf{Resampling} & cal500 & chess & corel5k & emotions & genbase & medical & scene & yeast \\ 
		\midrule
		BPMLL & None & 0.456$\pm$0.014 & 0.023$\pm$0.000 & 0.026$\pm$0.001 & 0.683$\pm$0.012 & 0.051$\pm$0.047 & 0.053$\pm$0.031 & 0.528$\pm$0.033 & 0.641$\pm$0.018 \\ 
		& LPROS & 0.461$\pm$0.019 & 0.022$\pm$0.001 & 0.025$\pm$0.000 & 0.668$\pm$0.037 & 0.052$\pm$0.048 & 0.033$\pm$0.030 & 0.482$\pm$0.021 & 0.641$\pm$0.017 \\ 
		& MLDM & \textbf{0.468$\pm$0.014} & 0.023$\pm$0.001 & 0.026$\pm$0.000 & 0.677$\pm$0.026 & 0.096$\pm$0.014 & 0.024$\pm$0.033 & \textbf{0.544$\pm$0.041} & \textbf{0.647$\pm$0.024} \\ 
		& MLROS & 0.459$\pm$0.018 & 0.023$\pm$0.001 & 0.028$\pm$0.001 & 0.687$\pm$0.018 & \textbf{0.106$\pm$0.017} & 0.052$\pm$0.030 & \textbf{0.544$\pm$0.033} & 0.638$\pm$0.016 \\ 
		& MLSMOTE & 0.462$\pm$0.020 & 0.023$\pm$0.001 & 0.021$\pm$0.000 & 0.662$\pm$0.038 & 0.088$\pm$0.021 & 0.035$\pm$0.032 & 0.517$\pm$0.048 & 0.633$\pm$0.011 \\ 
		& MLSOL & 0.457$\pm$0.013 & 0.023$\pm$0.001 & \textbf{0.029$\pm$0.001} & \textbf{0.688$\pm$0.041} & 0.074$\pm$0.047 & 0.046$\pm$0.042 & 0.515$\pm$0.016 & 0.641$\pm$0.015 \\ 
		& REMEDIAL & 0.417$\pm$0.011 & 0.020$\pm$0.001 & 0.021$\pm$0.000 & 0.621$\pm$0.023 & 0.072$\pm$0.041 & \textbf{0.064$\pm$0.004} & 0.520$\pm$0.018 & 0.614$\pm$0.009 \\ 
		\cmidrule{1-2}
		BR-J48 & None & 0.348$\pm$0.016 & 0.297$\pm$0.017 & 0.113$\pm$0.011 & 0.600$\pm$0.004 & 0.977$\pm$0.033 & \textbf{0.807$\pm$0.012} & \textbf{0.628$\pm$0.012} & 0.576$\pm$0.017 \\ 
		& LPROS & 0.348$\pm$0.016 & 0.284$\pm$0.028 & \textbf{0.126$\pm$0.009} & \textbf{0.592$\pm$0.039} & 0.943$\pm$0.081 & 0.805$\pm$0.015 & 0.618$\pm$0.008 & 0.551$\pm$0.009 \\ 
		& MLDM & 0.337$\pm$0.024 & 0.292$\pm$0.021 & 0.110$\pm$0.011 & \textbf{0.592$\pm$0.013} & \textbf{0.988$\pm$0.011} & 0.803$\pm$0.011 & 0.599$\pm$0.016 & \textbf{0.587$\pm$0.014} \\ 
		& MLROS & 0.342$\pm$0.010 & \textbf{0.298$\pm$0.017} & 0.123$\pm$0.007 & 0.567$\pm$0.027 & 0.987$\pm$0.038 & 0.795$\pm$0.018 & 0.624$\pm$0.029 & 0.549$\pm$0.021 \\ 
		& MLSMOTE & 0.318$\pm$0.008 & 0.282$\pm$0.020 & 0.109$\pm$0.008 & 0.575$\pm$0.014 & 0.927$\pm$0.080 & 0.798$\pm$0.008 & 0.618$\pm$0.019 & 0.550$\pm$0.020 \\ 
		& MLSOL & \textbf{0.363$\pm$0.014} & 0.291$\pm$0.018 & 0.122$\pm$0.010 & 0.579$\pm$0.020 & 0.932$\pm$0.082 & 0.802$\pm$0.014 & 0.616$\pm$0.011 & 0.560$\pm$0.016 \\ 
		& REMEDIAL & 0.155$\pm$0.023 & 0.105$\pm$0.032 & 0.049$\pm$0.007 & 0.417$\pm$0.033 & 0.935$\pm$0.083 & 0.734$\pm$0.015 & 0.613$\pm$0.025 & 0.522$\pm$0.057 \\ 
		\cmidrule{1-2}
		HOMER & None & \textbf{0.398$\pm$0.011} & 0.276$\pm$0.015 & 0.165$\pm$0.010 & \textbf{0.599$\pm$0.012} & 0.979$\pm$0.028 & \textbf{0.804$\pm$0.012} & 0.603$\pm$0.016 & 0.577$\pm$0.027 \\ 
		& LPROS & 0.396$\pm$0.016 & 0.268$\pm$0.006 & 0.159$\pm$0.005 & 0.586$\pm$0.019 & 0.947$\pm$0.077 & 0.779$\pm$0.010 & 0.608$\pm$0.011 & 0.553$\pm$0.015 \\ 
		& MLDM & 0.393$\pm$0.015 & 0.286$\pm$0.011 & 0.172$\pm$0.007 & 0.583$\pm$0.024 & 0.985$\pm$0.010 & \textbf{0.804$\pm$0.012} & 0.592$\pm$0.017 & \textbf{0.583$\pm$0.024} \\ 
		& MLROS & 0.353$\pm$0.019 & 0.267$\pm$0.016 & \textbf{0.173$\pm$0.009} & 0.577$\pm$0.022 & \textbf{0.987$\pm$0.014} & 0.779$\pm$0.012 & 0.592$\pm$0.016 & 0.553$\pm$0.016 \\ 
		& MLSMOTE & 0.340$\pm$0.013 & 0.286$\pm$0.011 & 0.164$\pm$0.005 & 0.566$\pm$0.013 & 0.923$\pm$0.092 & 0.795$\pm$0.009 & 0.598$\pm$0.021 & 0.559$\pm$0.009 \\ 
		& MLSOL & 0.387$\pm$0.009 & \textbf{0.302$\pm$0.008} & 0.163$\pm$0.005 & 0.592$\pm$0.030 & 0.930$\pm$0.088 & 0.784$\pm$0.013 & \textbf{0.610$\pm$0.010} & 0.554$\pm$0.021 \\ 
		& REMEDIAL & 0.394$\pm$0.011 & 0.248$\pm$0.017 & 0.138$\pm$0.007 & 0.502$\pm$0.050 & 0.944$\pm$0.085 & 0.760$\pm$0.039 & 0.601$\pm$0.022 & 0.581$\pm$0.037 \\ 
		\cmidrule{1-2}
		LP-J48 & None & 0.332$\pm$0.015 & 0.145$\pm$0.011 & 0.104$\pm$0.009 & 0.564$\pm$0.032 & 0.965$\pm$0.035 & \textbf{0.760$\pm$0.034} & 0.583$\pm$0.035 & 0.530$\pm$0.016 \\ 
		& LPROS & 0.332$\pm$0.015 & 0.152$\pm$0.015 & \textbf{0.109$\pm$0.006} & \textbf{0.583$\pm$0.009} & 0.931$\pm$0.099 & 0.743$\pm$0.021 & 0.575$\pm$0.021 & 0.524$\pm$0.017 \\ 
		& MLDM & 0.320$\pm$0.014 & 0.156$\pm$0.015 & \textbf{0.109$\pm$0.008} & 0.561$\pm$0.020 & \textbf{0.977$\pm$0.011} & 0.755$\pm$0.035 & 0.578$\pm$0.029 & 0.529$\pm$0.005 \\ 
		& MLROS & 0.334$\pm$0.018 & \textbf{0.160$\pm$0.015} & 0.105$\pm$0.002 & 0.566$\pm$0.019 & \textbf{0.977$\pm$0.042} & 0.748$\pm$0.026 & 0.587$\pm$0.017 & \textbf{0.540$\pm$0.015} \\ 
		& MLSMOTE & 0.317$\pm$0.009 & 0.150$\pm$0.009 & 0.106$\pm$0.006 & 0.554$\pm$0.024 & 0.919$\pm$0.108 & 0.747$\pm$0.036 & \textbf{0.596$\pm$0.020} & 0.525$\pm$0.013 \\ 
		& MLSOL & \textbf{0.339$\pm$0.012} & 0.158$\pm$0.013 & 0.108$\pm$0.005 & 0.559$\pm$0.025 & 0.905$\pm$0.117 & 0.756$\pm$0.012 & 0.594$\pm$0.018 & 0.524$\pm$0.017 \\ 
		& REMEDIAL & 0.273$\pm$0.018 & 0.048$\pm$0.019 & 0.043$\pm$0.009 & 0.530$\pm$0.024 & 0.918$\pm$0.111 & 0.685$\pm$0.054 & 0.575$\pm$0.020 & 0.427$\pm$0.034 \\ 
		\cmidrule{1-2}
		MLkNN & None & 0.316$\pm$0.010 & 0.063$\pm$0.024 & 0.027$\pm$0.005 & 0.676$\pm$0.016 & 0.941$\pm$0.053 & 0.665$\pm$0.017 & 0.730$\pm$0.012 & 0.640$\pm$0.023 \\ 
		& LPROS & 0.316$\pm$0.010 & \textbf{0.087$\pm$0.025} & 0.033$\pm$0.007 & 0.632$\pm$0.026 & 0.892$\pm$0.126 & 0.645$\pm$0.037 & 0.711$\pm$0.008 & 0.626$\pm$0.025 \\ 
		& MLDM & 0.322$\pm$0.013 & 0.071$\pm$0.032 & 0.027$\pm$0.004 & \textbf{0.680$\pm$0.010} & 0.946$\pm$0.009 & 0.665$\pm$0.017 & 0.721$\pm$0.025 & \textbf{0.642$\pm$0.026} \\ 
		& MLROS & 0.325$\pm$0.009 & 0.065$\pm$0.016 & 0.044$\pm$0.008 & 0.647$\pm$0.015 & \textbf{0.950$\pm$0.035} & 0.605$\pm$0.034 & 0.721$\pm$0.011 & 0.587$\pm$0.020 \\ 
		& MLSMOTE & 0.279$\pm$0.006 & 0.042$\pm$0.015 & 0.029$\pm$0.005 & 0.652$\pm$0.020 & 0.870$\pm$0.140 & 0.664$\pm$0.016 & 0.731$\pm$0.019 & 0.623$\pm$0.016 \\ 
		& MLSOL & \textbf{0.338$\pm$0.013} & 0.080$\pm$0.016 & \textbf{0.054$\pm$0.009} & 0.674$\pm$0.004 & 0.867$\pm$0.150 & \textbf{0.679$\pm$0.027} & \textbf{0.733$\pm$0.014} & 0.638$\pm$0.017 \\ 
		& REMEDIAL & 0.085$\pm$0.010 & 0.005$\pm$0.006 & 0.019$\pm$0.004 & 0.440$\pm$0.022 & 0.886$\pm$0.126 & 0.556$\pm$0.060 & 0.716$\pm$0.023 & 0.522$\pm$0.050 \\ 
		\bottomrule
	\end{tabular}
	\caption{Average classification performance as measured by the label-based micro-averaged F1 metric (the higher the better). The best values for each classifier/dataset quadrant are highlighted in bold.} 
\end{table*}

% latex table generated in R 4.2.2 by xtable 1.8-4 package
% Tue Jul  2 12:12:53 2024
\begin{table*}[h!]
	\centering\renewcommand{\arraystretch}{0.9}
	\tiny
	\begin{tabular}{llcccccccc}
		\toprule
		\textbf{Classifier} & \textbf{Resampling} & cal500 & chess & corel5k & emotions & genbase & medical & scene & yeast \\ 
		\midrule
		BPMLL & None & 0.252$\pm$0.013 & 0.575$\pm$0.047 & 0.565$\pm$0.015 & 0.211$\pm$0.007 & \textbf{0.424$\pm$0.433} & 0.629$\pm$0.355 & 0.243$\pm$0.060 & 0.228$\pm$0.010 \\ 
		& LPROS & 0.245$\pm$0.008 & 0.548$\pm$0.133 & 0.583$\pm$0.033 & 0.223$\pm$0.024 & 0.511$\pm$0.436 & 0.577$\pm$0.503 & 0.308$\pm$0.049 & 0.235$\pm$0.013 \\ 
		& MLDM & 0.245$\pm$0.003 & 0.616$\pm$0.027 & 0.571$\pm$0.020 & 0.214$\pm$0.012 & 0.621$\pm$0.281 & \textbf{0.357$\pm$0.451} & 0.243$\pm$0.044 & \textbf{0.224$\pm$0.013} \\ 
		& MLROS & 0.243$\pm$0.008 & 0.512$\pm$0.128 & 0.534$\pm$0.017 & \textbf{0.204$\pm$0.006} & 0.723$\pm$0.262 & 0.638$\pm$0.359 & \textbf{0.230$\pm$0.033} & 0.225$\pm$0.010 \\ 
		& MLSMOTE & \textbf{0.203$\pm$0.012} & 0.488$\pm$0.223 & 0.562$\pm$0.120 & 0.221$\pm$0.021 & 0.724$\pm$0.283 & 0.531$\pm$0.460 & 0.240$\pm$0.050 & 0.232$\pm$0.008 \\ 
		& MLSOL & 0.256$\pm$0.010 & 0.613$\pm$0.036 & 0.525$\pm$0.043 & 0.208$\pm$0.023 & 0.666$\pm$0.361 & 0.400$\pm$0.340 & 0.244$\pm$0.042 & 0.235$\pm$0.011 \\ 
		& REMEDIAL & 0.220$\pm$0.007 & \textbf{0.310$\pm$0.370} & \textbf{0.267$\pm$0.059} & 0.228$\pm$0.014 & 0.645$\pm$0.385 & 0.804$\pm$0.044 & 0.280$\pm$0.031 & 0.233$\pm$0.012 \\ 
		\cmidrule{1-2}
		BR-J48 & None & 0.165$\pm$0.005 & 0.011$\pm$0.000 & 0.010$\pm$0.000 & \textbf{0.248$\pm$0.010} & 0.002$\pm$0.003 & 0.011$\pm$0.001 & 0.133$\pm$0.004 & 0.251$\pm$0.010 \\ 
		& LPROS & 0.165$\pm$0.005 & 0.012$\pm$0.000 & 0.011$\pm$0.000 & 0.256$\pm$0.024 & 0.004$\pm$0.004 & 0.011$\pm$0.001 & 0.136$\pm$0.002 & 0.272$\pm$0.007 \\ 
		& MLDM & 0.167$\pm$0.005 & 0.011$\pm$0.000 & 0.010$\pm$0.000 & 0.258$\pm$0.010 & \textbf{0.001$\pm$0.001} & 0.011$\pm$0.001 & 0.144$\pm$0.006 & 0.245$\pm$0.010 \\ 
		& MLROS & 0.189$\pm$0.002 & 0.011$\pm$0.000 & 0.010$\pm$0.000 & 0.270$\pm$0.011 & \textbf{0.001$\pm$0.003} & 0.011$\pm$0.001 & 0.135$\pm$0.008 & 0.270$\pm$0.011 \\ 
		& MLSMOTE & 0.170$\pm$0.003 & 0.011$\pm$0.000 & 0.010$\pm$0.000 & 0.257$\pm$0.010 & 0.005$\pm$0.005 & 0.011$\pm$0.000 & 0.135$\pm$0.006 & 0.266$\pm$0.012 \\ 
		& MLSOL & 0.174$\pm$0.003 & 0.011$\pm$0.000 & 0.010$\pm$0.000 & 0.263$\pm$0.013 & 0.005$\pm$0.005 & 0.011$\pm$0.001 & 0.140$\pm$0.005 & 0.270$\pm$0.011 \\ 
		& REMEDIAL & \textbf{0.150$\pm$0.004} & 0.011$\pm$0.000 & \textbf{0.009$\pm$0.000} & 0.268$\pm$0.013 & 0.004$\pm$0.005 & 0.013$\pm$0.000 & \textbf{0.131$\pm$0.006} & \textbf{0.237$\pm$0.013} \\ 
		\cmidrule{1-2}
		HOMER & None & 0.186$\pm$0.011 & 0.014$\pm$0.000 & 0.013$\pm$0.000 & \textbf{0.252$\pm$0.007} & 0.002$\pm$0.002 & \textbf{0.011$\pm$0.001} & 0.142$\pm$0.005 & 0.263$\pm$0.014 \\ 
		& LPROS & 0.185$\pm$0.004 & 0.015$\pm$0.000 & 0.014$\pm$0.000 & 0.253$\pm$0.012 & 0.003$\pm$0.004 & 0.012$\pm$0.001 & 0.140$\pm$0.004 & 0.274$\pm$0.011 \\ 
		& MLDM & 0.195$\pm$0.006 & 0.014$\pm$0.001 & 0.013$\pm$0.000 & 0.262$\pm$0.015 & \textbf{0.001$\pm$0.001} & \textbf{0.011$\pm$0.001} & 0.149$\pm$0.007 & 0.259$\pm$0.017 \\ 
		& MLROS & 0.197$\pm$0.004 & 0.014$\pm$0.000 & 0.014$\pm$0.000 & 0.263$\pm$0.011 & \textbf{0.001$\pm$0.001} & 0.012$\pm$0.001 & 0.144$\pm$0.007 & 0.271$\pm$0.010 \\ 
		& MLSMOTE & \textbf{0.178$\pm$0.003} & 0.013$\pm$0.000 & 0.013$\pm$0.000 & 0.265$\pm$0.012 & 0.005$\pm$0.005 & \textbf{0.011$\pm$0.001} & 0.143$\pm$0.008 & 0.272$\pm$0.009 \\ 
		& MLSOL & 0.191$\pm$0.004 & 0.013$\pm$0.001 & 0.013$\pm$0.000 & 0.258$\pm$0.023 & 0.005$\pm$0.005 & 0.012$\pm$0.001 & 0.141$\pm$0.003 & 0.278$\pm$0.011 \\ 
		& REMEDIAL & 0.182$\pm$0.005 & \textbf{0.012$\pm$0.001} & \textbf{0.012$\pm$0.000} & 0.262$\pm$0.018 & 0.004$\pm$0.004 & 0.012$\pm$0.002 & \textbf{0.136$\pm$0.009} & \textbf{0.246$\pm$0.012} \\ 
		\cmidrule{1-2}
		LP-J48 & None & 0.198$\pm$0.004 & 0.018$\pm$0.001 & 0.017$\pm$0.000 & 0.273$\pm$0.019 & 0.003$\pm$0.003 & \textbf{0.013$\pm$0.002} & 0.150$\pm$0.014 & 0.284$\pm$0.009 \\ 
		& LPROS & 0.198$\pm$0.004 & 0.018$\pm$0.000 & 0.017$\pm$0.000 & 0.261$\pm$0.008 & 0.004$\pm$0.005 & 0.014$\pm$0.001 & 0.154$\pm$0.007 & 0.293$\pm$0.009 \\ 
		& MLDM & 0.214$\pm$0.013 & 0.018$\pm$0.001 & 0.017$\pm$0.000 & 0.276$\pm$0.011 & \textbf{0.002$\pm$0.001} & \textbf{0.013$\pm$0.002} & 0.152$\pm$0.011 & 0.285$\pm$0.001 \\ 
		& MLROS & 0.200$\pm$0.005 & 0.017$\pm$0.001 & 0.017$\pm$0.000 & 0.269$\pm$0.014 & \textbf{0.002$\pm$0.003} & 0.014$\pm$0.002 & 0.148$\pm$0.008 & \textbf{0.278$\pm$0.011} \\ 
		& MLSMOTE & \textbf{0.180$\pm$0.002} & 0.017$\pm$0.000 & 0.017$\pm$0.000 & 0.271$\pm$0.017 & 0.005$\pm$0.005 & 0.014$\pm$0.002 & \textbf{0.143$\pm$0.007} & 0.289$\pm$0.009 \\ 
		& MLSOL & 0.199$\pm$0.005 & 0.017$\pm$0.000 & 0.016$\pm$0.000 & 0.278$\pm$0.015 & 0.006$\pm$0.007 & \textbf{0.013$\pm$0.001} & 0.151$\pm$0.007 & 0.295$\pm$0.010 \\ 
		& REMEDIAL & 0.184$\pm$0.004 & \textbf{0.012$\pm$0.000} & \textbf{0.010$\pm$0.000} & \textbf{0.258$\pm$0.014} & 0.005$\pm$0.006 & 0.016$\pm$0.002 & 0.147$\pm$0.006 & 0.281$\pm$0.012 \\ 
		\cmidrule{1-2}
		MLkNN & None & \textbf{0.140$\pm$0.002} & 0.011$\pm$0.000 & \textbf{0.009$\pm$0.000} & 0.188$\pm$0.008 & 0.005$\pm$0.004 & \textbf{0.016$\pm$0.001} & 0.088$\pm$0.004 & \textbf{0.196$\pm$0.012} \\ 
		& LPROS & \textbf{0.140$\pm$0.002} & 0.011$\pm$0.000 & 0.010$\pm$0.000 & 0.213$\pm$0.012 & 0.007$\pm$0.006 & 0.017$\pm$0.001 & 0.100$\pm$0.005 & 0.209$\pm$0.013 \\ 
		& MLDM & 0.141$\pm$0.003 & 0.011$\pm$0.000 & \textbf{0.009$\pm$0.000} & \textbf{0.186$\pm$0.008} & 0.005$\pm$0.001 & \textbf{0.016$\pm$0.001} & 0.089$\pm$0.007 & \textbf{0.196$\pm$0.012} \\ 
		& MLROS & 0.141$\pm$0.002 & 0.011$\pm$0.000 & 0.010$\pm$0.000 & 0.205$\pm$0.004 & \textbf{0.004$\pm$0.003} & 0.018$\pm$0.002 & 0.092$\pm$0.005 & 0.213$\pm$0.010 \\ 
		& MLSMOTE & 0.142$\pm$0.002 & 0.011$\pm$0.000 & \textbf{0.009$\pm$0.000} & 0.204$\pm$0.011 & 0.008$\pm$0.006 & \textbf{0.016$\pm$0.001} & \textbf{0.088$\pm$0.005} & 0.205$\pm$0.008 \\ 
		& MLSOL & 0.142$\pm$0.002 & 0.011$\pm$0.000 & 0.010$\pm$0.000 & 0.194$\pm$0.006 & 0.008$\pm$0.007 & \textbf{0.016$\pm$0.001} & 0.093$\pm$0.004 & 0.205$\pm$0.009 \\ 
		& REMEDIAL & 0.146$\pm$0.003 & 0.011$\pm$0.000 & \textbf{0.009$\pm$0.000} & 0.255$\pm$0.004 & 0.007$\pm$0.005 & 0.018$\pm$0.002 & 0.090$\pm$0.006 & 0.220$\pm$0.013 \\ 
		\bottomrule
	\end{tabular}
	\caption{Average classification performance as measured by the sample-based Hamming loss metric (the lower the better). The best values for each classifier/dataset quadrant are highlighted in bold.} 
\end{table*}

% latex table generated in R 4.2.2 by xtable 1.8-4 package
% Tue Jul  2 12:12:53 2024
\begin{table*}[h!]
	\centering\renewcommand{\arraystretch}{0.9}
	\tiny
	\begin{tabular}{llcccccccc}
		\toprule
		\textbf{Classifier} & \textbf{Resampling} & cal500 & chess & corel5k & emotions & genbase & medical & scene & yeast \\ 
		\midrule
		BPMLL & None & 0.135$\pm$0.051 & 0.996$\pm$0.005 & \textbf{0.996$\pm$0.003} & 0.297$\pm$0.041 & 0.988$\pm$0.026 & 0.976$\pm$0.020 & 0.514$\pm$0.126 & 0.253$\pm$0.010 \\ 
		& LPROS & 0.146$\pm$0.059 & 0.996$\pm$0.005 & \textbf{0.996$\pm$0.002} & 0.305$\pm$0.021 & 0.993$\pm$0.008 & 0.969$\pm$0.016 & 0.575$\pm$0.044 & 0.267$\pm$0.017 \\ 
		& MLDM & 0.146$\pm$0.048 & 0.998$\pm$0.002 & \textbf{0.996$\pm$0.004} & 0.320$\pm$0.052 & 0.997$\pm$0.004 & 0.969$\pm$0.018 & \textbf{0.495$\pm$0.140} & \textbf{0.248$\pm$0.021} \\ 
		& MLROS & 0.132$\pm$0.063 & \textbf{0.995$\pm$0.005} & 0.997$\pm$0.002 & \textbf{0.294$\pm$0.025} & 0.985$\pm$0.025 & 0.974$\pm$0.023 & 0.513$\pm$0.057 & 0.273$\pm$0.027 \\ 
		& MLSMOTE & 0.171$\pm$0.056 & \textbf{0.995$\pm$0.006} & 0.999$\pm$0.002 & 0.332$\pm$0.053 & 0.992$\pm$0.011 & 0.990$\pm$0.009 & 0.559$\pm$0.112 & 0.273$\pm$0.027 \\ 
		& MLSOL & \textbf{0.123$\pm$0.048} & 0.998$\pm$0.003 & 0.998$\pm$0.002 & 0.299$\pm$0.035 & \textbf{0.978$\pm$0.025} & 0.977$\pm$0.021 & 0.524$\pm$0.061 & 0.258$\pm$0.015 \\ 
		& REMEDIAL & 0.220$\pm$0.051 & 0.998$\pm$0.002 & 0.999$\pm$0.001 & 0.313$\pm$0.042 & 0.992$\pm$0.007 & \textbf{0.960$\pm$0.023} & 0.589$\pm$0.033 & 0.285$\pm$0.015 \\ 
		\cmidrule{1-2}
		BR-J48 & None & 0.691$\pm$0.089 & \textbf{0.553$\pm$0.031} & \textbf{0.711$\pm$0.016} & \textbf{0.395$\pm$0.042} & \textbf{0.004$\pm$0.007} & \textbf{0.183$\pm$0.022} & 0.406$\pm$0.020 & 0.425$\pm$0.068 \\ 
		& LPROS & 0.691$\pm$0.089 & 0.584$\pm$0.012 & 0.755$\pm$0.005 & 0.399$\pm$0.049 & 0.042$\pm$0.086 & 0.201$\pm$0.018 & 0.422$\pm$0.012 & \textbf{0.344$\pm$0.035} \\ 
		& MLDM & 0.731$\pm$0.045 & 0.568$\pm$0.019 & 0.717$\pm$0.013 & 0.422$\pm$0.028 & 0.006$\pm$0.006 & 0.185$\pm$0.027 & 0.438$\pm$0.024 & 0.377$\pm$0.035 \\ 
		& MLROS & 0.826$\pm$0.061 & 0.574$\pm$0.026 & 0.725$\pm$0.016 & 0.427$\pm$0.027 & \textbf{0.004$\pm$0.003} & 0.205$\pm$0.040 & 0.413$\pm$0.023 & 0.355$\pm$0.036 \\ 
		& MLSMOTE & 0.707$\pm$0.024 & 0.561$\pm$0.024 & 0.713$\pm$0.014 & 0.396$\pm$0.060 & 0.046$\pm$0.083 & 0.202$\pm$0.013 & 0.401$\pm$0.024 & 0.384$\pm$0.039 \\ 
		& MLSOL & 0.745$\pm$0.043 & 0.575$\pm$0.020 & 0.721$\pm$0.014 & 0.452$\pm$0.050 & 0.047$\pm$0.089 & 0.185$\pm$0.031 & 0.425$\pm$0.015 & 0.380$\pm$0.057 \\ 
		& REMEDIAL & \textbf{0.679$\pm$0.060} & 0.584$\pm$0.008 & 0.712$\pm$0.005 & \textbf{0.395$\pm$0.031} & 0.048$\pm$0.096 & \textbf{0.183$\pm$0.019} & \textbf{0.391$\pm$0.021} & 0.400$\pm$0.024 \\ 
		\cmidrule{1-2}
		HOMER & None & 0.806$\pm$0.065 & 0.702$\pm$0.031 & \textbf{0.797$\pm$0.018} & \textbf{0.403$\pm$0.032} & \textbf{0.007$\pm$0.007} & \textbf{0.202$\pm$0.018} & 0.435$\pm$0.016 & 0.410$\pm$0.058 \\ 
		& LPROS & 0.812$\pm$0.043 & 0.715$\pm$0.018 & 0.825$\pm$0.013 & 0.427$\pm$0.035 & 0.047$\pm$0.089 & 0.256$\pm$0.031 & 0.436$\pm$0.013 & 0.368$\pm$0.044 \\ 
		& MLDM & 0.857$\pm$0.039 & \textbf{0.674$\pm$0.019} & 0.799$\pm$0.018 & 0.430$\pm$0.028 & 0.012$\pm$0.008 & 0.204$\pm$0.024 & 0.447$\pm$0.019 & 0.454$\pm$0.053 \\ 
		& MLROS & 0.842$\pm$0.052 & 0.697$\pm$0.041 & 0.801$\pm$0.019 & \textbf{0.403$\pm$0.020} & 0.010$\pm$0.006 & 0.230$\pm$0.013 & 0.466$\pm$0.016 & 0.390$\pm$0.034 \\ 
		& MLSMOTE & 0.789$\pm$0.045 & 0.682$\pm$0.024 & 0.805$\pm$0.009 & 0.432$\pm$0.045 & 0.067$\pm$0.133 & 0.216$\pm$0.011 & 0.450$\pm$0.036 & \textbf{0.357$\pm$0.071} \\ 
		& MLSOL & 0.847$\pm$0.020 & 0.675$\pm$0.026 & 0.799$\pm$0.011 & 0.406$\pm$0.019 & 0.054$\pm$0.096 & 0.233$\pm$0.022 & 0.430$\pm$0.018 & 0.405$\pm$0.044 \\ 
		& REMEDIAL & \textbf{0.745$\pm$0.034} & 0.724$\pm$0.021 & 0.830$\pm$0.010 & 0.436$\pm$0.046 & 0.056$\pm$0.100 & 0.221$\pm$0.063 & \textbf{0.420$\pm$0.029} & 0.395$\pm$0.104 \\ 
		\cmidrule{1-2}
		LP-J48 & None & 0.988$\pm$0.008 & 0.954$\pm$0.021 & 0.988$\pm$0.003 & 0.445$\pm$0.034 & 0.028$\pm$0.037 & 0.231$\pm$0.035 & 0.414$\pm$0.029 & 0.538$\pm$0.031 \\ 
		& LPROS & 0.988$\pm$0.008 & 0.930$\pm$0.017 & \textbf{0.973$\pm$0.007} & \textbf{0.421$\pm$0.032} & 0.070$\pm$0.093 & 0.237$\pm$0.026 & 0.433$\pm$0.023 & \textbf{0.467$\pm$0.043} \\ 
		& MLDM & 0.988$\pm$0.008 & 0.940$\pm$0.014 & 0.984$\pm$0.007 & 0.450$\pm$0.028 & \textbf{0.010$\pm$0.008} & \textbf{0.230$\pm$0.035} & 0.424$\pm$0.022 & 0.507$\pm$0.015 \\ 
		& MLROS & \textbf{0.977$\pm$0.016} & 0.921$\pm$0.009 & 0.984$\pm$0.003 & 0.453$\pm$0.037 & \textbf{0.010$\pm$0.050} & 0.245$\pm$0.016 & 0.408$\pm$0.021 & 0.501$\pm$0.022 \\ 
		& MLSMOTE & \textbf{0.977$\pm$0.017} & 0.952$\pm$0.012 & 0.985$\pm$0.008 & 0.440$\pm$0.054 & 0.086$\pm$0.103 & 0.250$\pm$0.036 & \textbf{0.399$\pm$0.022} & 0.537$\pm$0.017 \\ 
		& MLSOL & 0.988$\pm$0.008 & \textbf{0.917$\pm$0.016} & 0.985$\pm$0.002 & 0.444$\pm$0.045 & 0.081$\pm$0.114 & 0.238$\pm$0.011 & 0.414$\pm$0.025 & 0.551$\pm$0.025 \\ 
		& REMEDIAL & 0.988$\pm$0.008 & 0.979$\pm$0.012 & 0.989$\pm$0.002 & 0.448$\pm$0.085 & 0.077$\pm$0.112 & 0.370$\pm$0.026 & 0.429$\pm$0.018 & 0.660$\pm$0.037 \\ 
		\cmidrule{1-2}
		MLkNN & None & \textbf{0.117$\pm$0.040} & \textbf{0.696$\pm$0.038} & 0.741$\pm$0.010 & \textbf{0.245$\pm$0.038} & \textbf{0.003$\pm$0.004} & \textbf{0.241$\pm$0.032} & \textbf{0.229$\pm$0.021} & 0.229$\pm$0.026 \\ 
		& LPROS & \textbf{0.117$\pm$0.040} & 0.730$\pm$0.028 & 0.800$\pm$0.012 & 0.279$\pm$0.015 & 0.064$\pm$0.107 & 0.264$\pm$0.043 & 0.241$\pm$0.022 & 0.237$\pm$0.029 \\ 
		& MLDM & 0.129$\pm$0.047 & 0.704$\pm$0.033 & 0.747$\pm$0.010 & 0.250$\pm$0.027 & 0.018$\pm$0.012 & 0.243$\pm$0.031 & \textbf{0.229$\pm$0.020} & \textbf{0.227$\pm$0.027} \\ 
		& MLROS & 0.139$\pm$0.065 & 0.746$\pm$0.019 & 0.756$\pm$0.012 & 0.284$\pm$0.034 & 0.012$\pm$0.008 & 0.293$\pm$0.035 & 0.242$\pm$0.018 & 0.237$\pm$0.035 \\ 
		& MLSMOTE & 0.182$\pm$0.042 & 0.704$\pm$0.035 & 0.748$\pm$0.013 & 0.272$\pm$0.054 & 0.063$\pm$0.121 & 0.254$\pm$0.035 & \textbf{0.229$\pm$0.014} & 0.241$\pm$0.026 \\ 
		& MLSOL & 0.127$\pm$0.056 & 0.711$\pm$0.033 & 0.748$\pm$0.021 & 0.289$\pm$0.046 & 0.083$\pm$0.136 & 0.253$\pm$0.046 & 0.238$\pm$0.020 & 0.247$\pm$0.019 \\ 
		& REMEDIAL & 0.175$\pm$0.032 & 0.715$\pm$0.040 & \textbf{0.734$\pm$0.016} & 0.336$\pm$0.045 & 0.061$\pm$0.117 & 0.254$\pm$0.042 & 0.233$\pm$0.019 & 0.238$\pm$0.028 \\ 
		\bottomrule
	\end{tabular}
	\caption{Average classification performance as measured by the ranking-based One error metric (the lower the better). The best values for each classifier/dataset quadrant are highlighted in bold.} 
\end{table*}	
	
\end{document}